\documentclass[10pt,journal,compsoc]{IEEEtran}

\ifCLASSOPTIONcompsoc
  \usepackage[nocompress]{cite}
\else
  \usepackage{cite}
\fi
\usepackage[switch]{lineno}
\usepackage{times}
\usepackage{epsfig}
\usepackage{graphicx}
\usepackage{amsmath}
\usepackage{amssymb}
\usepackage{multirow,makecell,footmisc,url}
\usepackage{subcaption}
\usepackage{gensymb}
\usepackage{color}

\newcommand{\yue}[1]{\textcolor[rgb]{0,0,0}{#1}}
\newcommand{\minor}[1]{\textcolor[rgb]{0,0,0}{#1}}

\newcommand{\jiarui}[1]{\textcolor[rgb]{0,0,0}{#1}}

\ifCLASSINFOpdf
\else
\fi

\begin{document}

\title{Global Context Networks}

\author{Yue Cao*,
        Jiarui Xu*,
        Stephen Lin,
        Fangyun Wei,
        Han Hu
\IEEEcompsocitemizethanks{\IEEEcompsocthanksitem * Equal contribution. \IEEEcompsocthanksitem Yue Cao, Stephen Lin, Fangyun Wei and Han Hu are with Microsoft Research Asia. Jiarui Xu is with Hong Kong University of Science and Technology. This work was done when Jiarui Xu was an intern at Microsoft Research Asia. E-mail: Yue Cao (yuecao@microsoft.com), Jiarui Xu (xvjiarui0826@gmail.com), Stephen Lin (stevelin@microsoft.com), Fangyun Wei (fawe@microsoft.com). Correspondence to: Han Hu (hanhu@microsoft.com). \IEEEcompsocthanksitem A preliminary version of this manuscript was published in \cite{cao2019gcnet}.}}

\markboth{IEEE Transactions on Pattern Analysis and Machine Intelligence}%
{Cao and Xu  \MakeLowercase{\textit{et al.}}: Global Context Networks}

\IEEEtitleabstractindextext{%
\begin{abstract}
The Non-Local Network (NLNet) presents a pioneering approach for capturing long-range dependencies within an image, via aggregating query-specific global context to each query position. However, through a rigorous empirical analysis, we have found that the global contexts modeled by the non-local network are almost the same for different query positions. In this paper, we take advantage of this finding to create a simplified network based on a query-independent formulation, which maintains the accuracy of NLNet but with significantly less computation. We further replace the one-layer transformation function of the non-local block by a two-layer bottleneck, which further reduces the parameter number considerably. The resulting network element, called the global context (GC) block, effectively models global context in a lightweight manner, allowing it to be applied at multiple layers of a backbone network to form a global context network (GCNet).  Experiments show that GCNet generally outperforms NLNet on major benchmarks for various recognition tasks. The code and network configurations are available at \url{https://github.com/xvjiarui/GCNet}.
\end{abstract}

}

\maketitle

\IEEEdisplaynontitleabstractindextext

\IEEEpeerreviewmaketitle

\IEEEraisesectionheading{\section{Introduction}\label{sec:introduction}}
Long-range dependencies among pixels in an image are essential to capture for global understanding of a visual scene. This dependency modeling is proven to benefit a wide range of recognition tasks, \minor{such as image classification \cite{hu2018senet}, object detection and segmentation~\cite{hu2018relation,zhang2018context}, and video action recognition~\cite{wang2017non}.} In convolutional neural networks, long-range dependencies are mainly modeled by deep stacking of convolution layers, where each layer models pixel relationships within a local neighborhood. However, direct repetition of convolution layers is computationally inefficient and hard to optimize \cite{wang2017non}, due in part to difficulties in delivering messages between distant positions.

To address this issue, the non-local network (NLNet) \cite{wang2017non} utilizes a layer to model long-range dependencies, via a self-attention mechanism~\cite{vaswani2017attention}. For each query position, the non-local network first computes pairwise relations between the query position and all other positions to form an attention map, and then aggregates the features of all positions by a weighted sum with the weights defined by the attention map. The aggregated features are finally added to the features of each query position to form the output.

\begin{figure}[]
    \includegraphics[width=1.0\columnwidth]{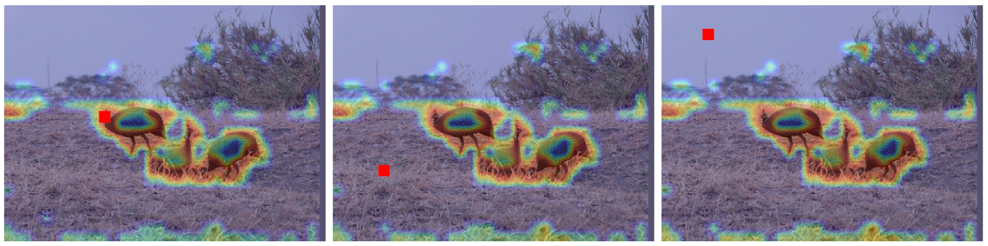}
	\caption{Visualization of attention maps (heatmaps) for different query positions (red points) in a non-local block on COCO object detection. The three attention maps are all almost the same. More examples are presented in Figure \ref{fig:vis-nlocal}.}
	\label{fig:vis-teaser}
\end{figure}

The query-specific attention weights in the non-local network are expected to reflect the importance of the corresponding positions to the query position. Visualizing these weights would help to better understand their behavior, but such analysis was largely missing in the original paper. In an analysis that we conducted, a surprising observation can be made. As shown in Figure~\ref{fig:vis-teaser}, we found that the attention maps for different query positions are almost the same, indicating that the learnt dependency is basically query-independent. This observation is further verified by the statistical analysis in Tables~\ref{table:statistical-nlocal}, \ref{table:statistical-nlocal-stages} and \ref{table:statistical-nlocal-finegrained}, which show that the distance between the attention maps of different query positions is very small.
This observation is verified in three standard tasks, object detection on COCO, image classification on ImageNet and action recognition on Kinetics.

Based on this observation, we propose a simplification of the non-local block in which a query-independent attention map is explicitly used for all query positions. The output is then formed by the same aggregation of features using this attention map as weights. This simplified block requires significantly less computation than the original non-local block, but exhibits almost no decrease in accuracy on several important visual recognition tasks. The block design follows a general three-step framework: (a) a {context modeling} module which aggregates the features of all positions together to form a global context feature; (b) a feature {transform} module to capture the channel-wise interdependencies; and (c) a {fusion} module to merge the global context feature into features of all positions. 
We further significantly reduce the parameter number by replacing the one-layer transformation function of the non-local block with a bottleneck of two layers, to form a new unit that we call the global context (GC) block.

Because of the lightweight computation of the GC block, it can be applied to all residual blocks in the ResNet architecture, in contrast to the original non-local block which is usually applied after just one or a few layers due to its heavy processing. We refer to this network as the global context network (GCNet). On COCO object detection/instance segmentation, it is found that GCNet outperforms NLNet by 1.9\% on AP${^\text{box}}$ and 1.5\% on AP${^\text{mask}}$ with just a 0.07\% relative increase in FLOPs. In addition, GCNet yields significant performance gains over four general visual recognition tasks: {object detection/segmentation on COCO} (2.7\%$\uparrow$ on AP$^\text{bbox}$, and 2.4\%$\uparrow$ on AP$^\text{mask}$ over Mask R-CNN with FPN and ResNet-50 as backbone \cite{he2017mask}), semantic segmentation on Cityscapes (3.2\%$\uparrow$ on mIoU over ResNet-101 as backbone with dilated convolutions), {image classification on ImageNet} (0.8\%$\uparrow$ on top-1 accuracy over ResNet-50 \cite{he2015resnet}), and {action recognition on Kinetics} (1.1\%$\uparrow$ on top-1 accuracy over the ResNet-50 Slow-only baseline \cite{feichtenhofer2018slowfast}), with less than a 0.26\% increase in computation cost.

\section{Related Work}
\subsection{Deep architectures}
Recent progress in computer vision have largely been driven by the improvement of basic deep architectures, which extract features for visual elements. One direction of improvement is to design better functional formulations of basic components to elevate the power of deep networks for the general purpose of image feature extraction.
A pioneering work along this path is AlexNet~\cite{krizhevsky2012imagenet}, which proves that increasing the depth and width of convolutional neural networks can achieve impressive accuracy in classifying objects in ImageNet. Since then, vast improvements have been made to unleash the power of deep architectures. VGG \cite{simonyan2014vgg} further increases the depth and width, and replaces most large-kernel convolution layers by smaller ones of $3\times 3$, which has become widely used in subsequent architecture designs.
GoogLeNet \cite{szegedy2015googlenet} extends the idea of the multi-branch layer from NIN \cite{lin2013nin} and introduces $1\times 1$ convolution to reduce the number of parameters.
ResNet \cite{he2015resnet} introduces skip connections (also called shortcuts), which can significantly reduce the gradient vanishing issue and allows the network to be tremendously deep.
In DenseNet \cite{huang2017densely}, every layer obtains additional inputs from all preceding layers and passes its own feature maps to all subsequent layers, through concatenation operations.
ResNeXt \cite{xie2017resnext} and Xception \cite{chollet2017xception} adopt group convolution to increase cardinality and reduce the redundancy of the network parameters.
Deformable Convolution Networks \cite{dai2017dcnv1,zhu2018dcnv2} present deformable convolutions for enhancing geometric modeling ability, which can significantly improve performance on fine-grained recognition tasks.
Local Relation Networks~\cite{hu2019localrelation} replace all spatial convolution layers by local relation layers, which adaptively determine aggregation weights based on the compositional relationship of local pixel pairs. Different from handcrafted architectures, automatic search of the cell structure for deep architectures has attracted much attention recently \cite{zoph2018nasnet,zhong2018practical}.

Another direction of improvement is to invent deep architectures for specific tasks, \minor{such as semantic segmentation~\cite{long2015fully,paszke2016enet,zhao2017pspnet,zhao2018icnet,chen2017rethinking,chen2018encoder}, object detection~\cite{dai2016r,redmon2016you,redmon2017yolo9000,lin2017focal,hu2018relation}, and video action recognition~\cite{xie2018rethinking,feichtenhofer2016convolutional,simonyan2014two,feichtenhofer2018slowfast}.}
MobileNet \cite{howard2017mobilenets,sandler2018mobilenetv2} is designed to adopt depthwise separable convolution as the basic block for mobile and embedded vision applications.
ShuffleNet \cite{zhang2018shufflenet,ma2018shufflenetv2} adopts channel shuffling, which facilitates the use of group convolution with 1x1 convolutions.
Fully-convolutional Networks (FCN) \cite{long2015fully,zhao2017pspnet,chen2017rethinking} are designed to make dense predictions for per-pixel tasks like semantic segmentation.
The YOLO series \cite{redmon2016you,redmon2017yolo9000} frames object detection as the regression of spatially separated bounding boxes and associated class probabilities, which is both fast and effective.
For video action recognition tasks, to better incorporate temporal information in feature extraction, I3D \cite{carreira2017i3d} introduces 3D convolution to deep networks.
To reduce the computation cost, P3D~\cite{qiu2017P3D} separates the 3D convolution into a sequence of temporal-only convolution and spatial-only convolution.

The proposed global context network is a new architecture designed for general purpose. It introduces a novel global context block which models long-range information into existing architectures, showing general improvements on a wide range of vision tasks, such as object detection, instance segmentation, image classification and action recognition.

\subsection{Long-range dependency modeling}

While existing deep architectures mainly work by stacking layers which operate locally, there are also methods that directly model long-range dependency using a single layer. Such methods can be categorized into two classes: pairwise based, and context fusion based.

Most pairwise methods are based on the self-attention mechanism, and the non-local network (NLNet) is a pioneering work~\cite{wang2017non} for pixel-pixel pairwise relation modeling that has proven beneficial for several visual recognition tasks, such as object detection and action recognition.
There are also extensions of non-local networks proposed to benefit specific tasks.
Object Context Networks (OCNet) \cite{yuan2018ocnet} model pixel-wise relationships in the same object category via self-attention mechanisms and also capture context at multiple scales.
Dual Attention Networks (DANet) \cite{fu2019danet} use self-attention mechanisms to model pixel-pixel relationships and channel-channel relationships to improve feature representations.
Criss-Cross Networks (CCNet) \cite{huang2018ccnet} accelerate NLNet via stacking two criss-cross blocks, which can enlarge the dependency range to the whole feature map with low computational cost.

While it is widely believed that NLNet benefits visual recognition due to pairwise relation modeling, this paper empirically proves that such belief is actually incorrect. In fact, for several important visual recognition tasks such as ImageNet image classification, COCO object detection and Kinetics action recognition, we observe that NLNet degenerates to learning the same global context vector for different pixels, and thus the effectiveness of NLNet can mainly be ascribed to global context modeling other than pairwise relation modeling. For some other visual recognition tasks, such as semantic segmentation, although we observe that some kind of pairwise relation is learnt, the accuracy improvement is still mostly ascribed to its global context modeling ability. Based on this observation, we propose a simplification of the non-local block, which explicitly learns global context other than pairwise relations. The resulting block, called the global context (GC) block, consumes significantly less computation than the non-local block but performs with the same accuracy on several important tasks. \yue{Note while the proposed GC block exploits the findings of this degeneration issue to explicitly simplify the non-local block, in a follow-up to this paper, our work on disentangled non-local networks (DNL) \cite{yin2020DNL} on the contrary attempts to alleviate this degeneration problem by a disentangled design in a manner that allows learning of different contexts for different pixels while preserving the shared global context.}

Different from pairwise methods, context fusion methods operate by strengthening the feature of each position by a context feature that aggregates information from all pixels including those at long range. For example, SENet \cite{hu2018senet} fuse the two features by adaptive rescaling on different channels.
GENet \cite{hu2018gather} uses local patches to compute position-adaptive context features.
PSANet \cite{zhao2018psanet} proposes to connect each position on the feature map to all the other ones through a self-adaptively learned attention mask, and aggregate the features of other positions via rescaling.
CBAM \cite{woo2018cbam} recalibrates the importance of both different spatial positions and channels also via rescaling.
All these methods adopt rescaling for feature aggregation, which may be of limited effectiveness for global context modeling.

The proposed GCNet is also a context fusion method. But by using a different context feature computation method (attention pooling) and a different fusion method (addition), GCNet performs generally better than the widely used SENet method. Noting that the context feature computation and fusion methods used in GCNet are inherited from NLNet, the proposed GCNet can be also seen as a product of connecting two representative long-range dependency modeling methods, NLNet and SENet, but makes good use of their respective strengths (GCNet is the same as NLNet in better context modeling and information fusion, while being as lightweight as SENet).

\begin{figure*}[!htb]
    \centering
    \includegraphics[width=1.0\linewidth]{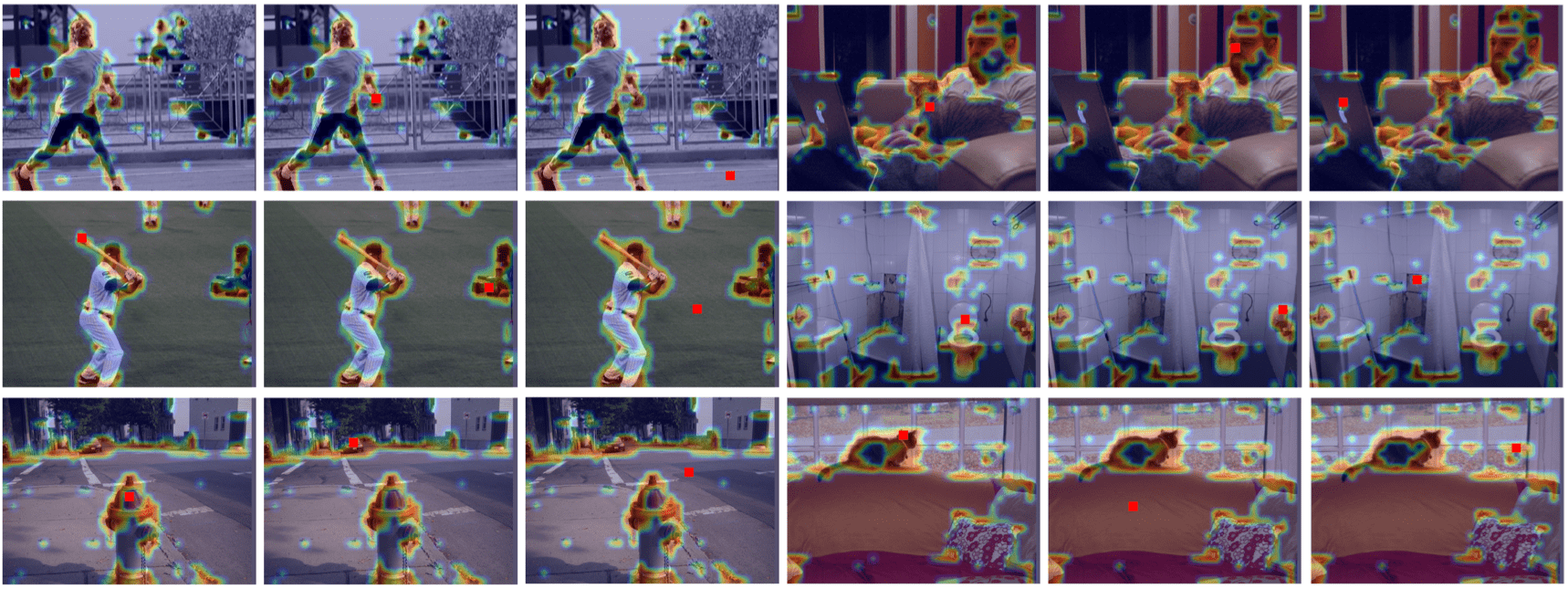}
	\caption{Visualization of attention maps (heatmaps) for different query positions (red points) in a non-local block on COCO object detection. For the same image, the attention maps of different query points are almost the same. \emph{Best viewed in color}.}
	\label{fig:vis-nlocal}
\end{figure*}

\subsection{Self-attention modeling} 

This paper is also related to the general self-attention mechanism whose application extends beyond pixel relation modeling \cite{gehring2016convolutional,chen2019graph,chiu2018state,devlin2018bert,zhu2019empirical,dong2019unified,garcia2017few,gehring2017convolutional,vaswani2017attention,velivckovic2017GAT,zhang2018SAGAN,wang2017residual,hu2018relation,wang2017non,yuan2018ocnet,hu2019localrelation,xu2019strn,chen20mega}.

In natural language processing, Transformer \cite{vaswani2017attention}, which applies a self-attention mechanism to model long-range dependencies between words, is a milestone work for machine translation.
Graph Attention Networks (GAT) \cite{velivckovic2017GAT} improve graph convolution with self-attention mechanisms that operate on graph-structured data, producing remarkable gains over baseline graph convolution methods.
Self-attention Generative Adversarial Networks (SAGAN) \cite{zhang2018SAGAN} generate high-resolution details as a function of not only spatially local points but also distant points, via self-attention mechanisms that model long-range dependency.

For visual recognition, aside from pixel relation modeling, the attention mechanism is also applied for object-object/object-pixel relation modeling~\cite{hu2018relation,gu2018learnregionfeat}, which is proven effective in object detection. 

The presented analysis and proposed GCNet in this paper are basically about the general self-attention mechanism, with experiments and instantiations mainly targeting the problem of pixel-pixel relation modeling. Such an analysis and global context modeling approach could be extended to other self-attention applications such as object-object/object-pixel relation modeling, natural language processing, and graph social networks. For these applications, there are questions of whether the pairwise relations can be well learnt by the self-attention mechanism and how the global context modeling approach can effectively contribute. Both of these questions on broader applications are promising directions for further study.

\section{Analysis of Non-local Networks}\label{sec:analysis}
In this section, we first review the design of the non-local block \cite{wang2017non}. While in-depth studies have been rare on what a non-local block learns and what makes it effective, we conduct such a study both qualitatively and statistically. Qualitatively, we visualize the attention maps across different query positions generated by a widely-used instantiation of the non-local block. 
Statistically, we compute the average cosine distances between different feature maps (including input, attention map, output and so on) inside the non-local block, to delve deep into the non-local block design. 
This in-depth study brings a new understanding of the non-local block and may inspire new approaches as in the next section. 

\begin{figure}[!htb]
    \centering
    \includegraphics[width=0.9\columnwidth]{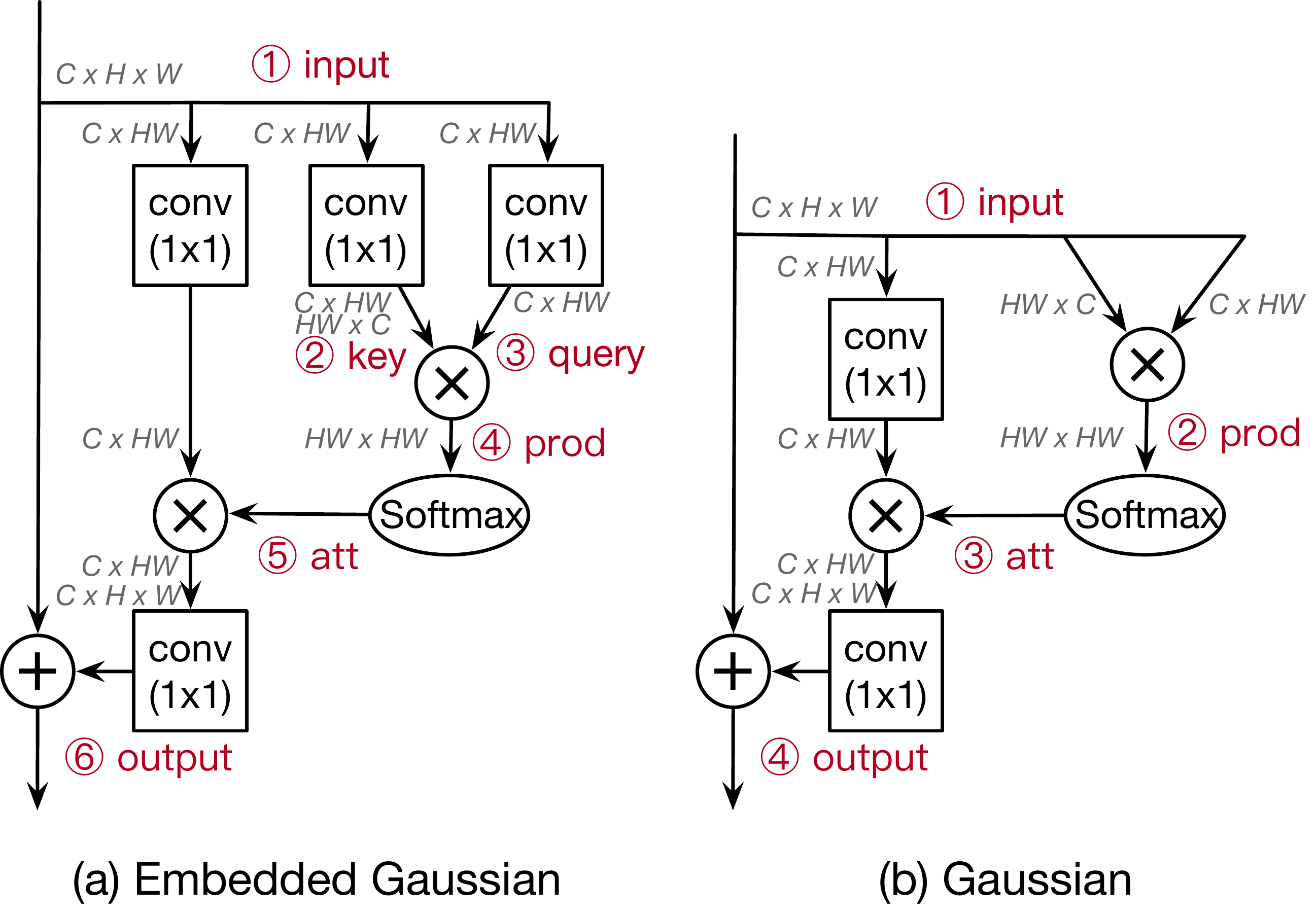}
	\caption{Two instantiations of the non-local block: Embedded Gaussian and Gaussian. The feature maps are shown by their dimensions, e.g. CxHxW. $\otimes$ denotes matrix multiplication, and $\oplus$ is broadcast element-wise addition. For two matrices with different dimensions, broadcast operations first broadcast features in each dimension to match the dimensions of the two matrices. The feature maps marked in red (e.g. `\textcircled{5} att') are statistically analyzed in Tables~\ref{table:statistical-nlocal}, \ref{table:statistical-nlocal-finegrained} and \ref{table:statistical-nlocal-stages}.}
	\label{fig:arch-nl}
\end{figure}

\subsection{Revisiting the Non-local Block} 
The basic non-local block \cite{wang2017non} aims at strengthening the features of the query position via aggregating information from other positions.
We denote ${\bf x}$=$\{{\bf x}_i\}_{i=1}^{N_p}$ as the feature map of one input instance (e.g., an image or video), where $N_p$ is the number of positions in the feature map (e.g., $N_p$=H$\cdot$W for image, $N_p$=H$\cdot$W$\cdot$T for video). ${\bf x}$ and ${\bf z}$ denote the input and output of the non-local block, respectively, which have the same dimensions. The non-local block is formulated as
\begin{equation}
    {{\bf{z}}_i} = {{\bf{x}}_i} + {W_z}\sum\nolimits_{j=1}^{N_p} {\frac{{f\left( {{{\bf{x}}_i},{{\bf{x}}_j}} \right)}}{{{\cal C}\left( {\bf{x}} \right)}}\left( {{W_v} \cdot {{\bf{x}}_j}} \right)},
\end{equation}
where $i$ is the index of query positions, and $j$ enumerates all possible positions. 
$f\left( {{{\bf{x}}_i},{{\bf{x}}_j}} \right)$ denotes the relationship between position $i$ and $j$, and has a normalization factor ${\cal C}\left( {\bf{x}} \right)$.
$W_z$ and $W_v$ denote linear transform matrices (e.g., 1x1 convolution).
For simplification, we denote $\omega_{ij}={\frac{{f\left( {{{\bf{x}}_i},{{\bf{x}}_j}} \right)}}{{{\cal C}\left( {\bf{x}} \right)}}}$ as the normalized pairwise relationship between position $i$ and $j$.

In \cite{wang2017non}, four instantiations of the non-local block are provided by defining $\omega_{ij}$ as different functions:
\begin{itemize}
	\item \emph{Gaussian}. $f$ in $\omega_{ij}$ is the Gaussian function, defined as ${\omega _{ij}}$=$\frac{{\exp \left( {\left\langle {{{\bf{x}}_i},{{\bf{x}}_j}} \right\rangle } \right)}}{{\sum\limits_m {\exp \left( {\left\langle {{{\bf{x}}_i},{{\bf{x}}_m}} \right\rangle } \right)} }}$;
	\item \emph{Embedded Gaussian}. It is a simple extension of Gaussian by using an embedding space to compute similarity, defined as ${\omega _{ij}}$=$\frac{{\exp \left( {\left\langle {{W_q}{{\bf{x}}_i},{W_k}{{\bf{x}}_j}} \right\rangle } \right)}}{{\sum\limits_m {\exp \left( {\left\langle {{W_q}{{\bf{x}}_i},{W_k}{{\bf{x}}_m}} \right\rangle } \right)} }}$;
	\item \emph{Dot product}. $f$ in $\omega_{ij}$ is defined as a dot-product similarity, formulated as ${\omega _{ij}}$=$\frac{{\left\langle {{W_q}{{\bf{x}}_i},{W_k}{{\bf{x}}_j}} \right\rangle }}{{{N_p}}}$;
	\item \emph{Concat}. It is defined as ${\omega _{ij}}$=$\frac{{{\rm{ReLU}}\left( {{W_q}\left[ {{{\bf{x}}_i},{{\bf{x}}_j}} \right]} \right)}}{{{N_p}}}$.
\end{itemize}
We illustrate the architecture of two most widely-used instantiations, Embedded Gaussian and Gaussian, in Figure \ref{fig:arch-nl}(a) and \ref{fig:arch-nl}(b).

The non-local block can be regarded as a query-specific global context modeling block, which strengthens the feature at a query position by a query-specific global context vector, computed by a weighted sum over all positions. The weights are determined by a similarity between two positions, and the weights over all positions form an attention map for one query position. 
The time and space complexity of the non-local block are heavy in that they are both quadratic to the number of positions $N_p$. Likely as a result, it is applied to only a few places in a network architecture, e.g. as one block inserted into the Mask R-CNN framework.

The non-local block \cite{wang2017non} is proven to benefit many visual recognition tasks, such as object detection/instance segmentation, and action recognition. It is believed that such effectiveness arises from effective learning of pairwise pixel relations \cite{wang2017non}. Nevertheless, direct evidence and an in-depth study of this has been lacking. In the following, we analyze what is truly learnt in non-local networks, both qualitatively and statistically. Such a study shed light on the behavior of non-local networks.

\subsection{Analysis}

\subsubsection{Visualization}
To intuitively understand the behavior of the non-local block, we first visualize the attention maps for different query positions.
As different instantiations achieve comparable performance \cite{wang2017non}, here we only visualize the most widely-used version, Embedded Gaussian, which has the same formulation as the block proposed in \cite{vaswani2017attention}.
Since attention maps in videos are hard to visualize and understand, we only show visualizations on object detection/instance segmentation, which takes images as input.
Following the standard setting of non-local networks for object detection \cite{wang2017non}, we conduct experiments on Mask R-CNN with FPN and ResNet50, and only add one non-local block right before the last residual block of res$_4$.

In Figure \ref{fig:vis-nlocal}, we randomly select six images from the COCO dataset, and visualize three different query positions (red points) and their query-specific attention maps (heatmaps) for each image. 
We surprisingly find that \textbf{for different query positions, their attention maps are almost the same}. This suggests that it may be redundant for the non-local block to compute different attention maps for different positions in object detection, as the non-local block may not learn pixel-pixel relationships in this task but rather just global context.
This observation motivates us to delve deep into the design of non-local block, to understand its real behavior.

\begin{table}[]
    \centering
    \addtolength{\tabcolsep}{-3pt}
    \small
\begin{tabular}{c|c|cc|ccc}
\Xhline{1.0pt}
\multirow{2}{*}{Dataset} & \multirow{2}{*}{Method} & \multirow{2}{*}{AP$^\text{bbox}$} & \multirow{2}{*}{AP$^\text{mask}$}  & \multicolumn{3}{c}{cosine distance} \\
\cline{5-7}
 &  &  &  & input & output  & att \\
\hline
\multirow{5}{35pt}{\centering COCO} 
 & \jiarui{baseline} &  37.2 & 33.8 & - & - & - \\
 & Gaussian &  38.0 & 34.8 & 0.397 & 0.062 & 0.177 \\
 & E-Gaussian & 38.0 & 34.7 & 0.402 & 0.012 & 0.020 \\
 & Dot product &  38.1 & 34.8 & 0.405 & 0.020 & 0.015 \\
 & Concat &  38.0 & 34.9 & 0.393 & 0.003 & 0.004 \\
 \hline
Dataset & Method & Top-1 & Top-5   & input & output  & att \\
 \hline
\multirow{5}{35pt}{\centering Kinetics} 
 & \jiarui{baseline} & 74.9  & 91.9 & - & - & - \\
 & Gaussian & 76.0 & 92.3 & 0.345 & 0.056 & 0.056 \\
 & E-Gaussian & 75.9 & 92.2 & 0.358 & 0.003 & 0.004 \\
 & Dot product & 76.0 & 92.3 & 0.353 & 0.095 & 0.099 \\
 & Concat & 75.4 & 92.2 & 0.354 & 0.048 & 0.049 \\
 \hline
Dataset & Method & Top-1 & Top-5   & input & output  & att \\
 \hline
\multirow{5}{35pt}{\centering ImageNet} 
 & \jiarui{baseline} & 76.5  & 93.4 & - & - & - \\
 & Gaussian & 77.1 & 93.6 & 0.045 & 0.005 & 0.011 \\
 & E-Gaussian & 77.2 & 91.9 & 0.301 & 0.074 & 0.115 \\
 & Dot product & 77.0 & 93.5 & 0.396 & 0.081 & 0.098 \\
 & Concat & 76.9 & 93.5 & 0.379 & 0.023 & 0.090 \\
\Xhline{1.0pt}
\end{tabular}
    \caption{Statistical analysis on four instantiations of non-local blocks. `input' denotes the input of the non-local block (${\bf x}_i$), `output' denotes the output of the non-local block (${\bf z}_i-{\bf x}_i$), `att' denotes the attention map of query positions ($\omega_i$).}
\label{table:statistical-nlocal}
\end{table}

\begin{table*}[t]
\begin{minipage}{0.45\linewidth}
\centering
    \addtolength{\tabcolsep}{-1.5pt}
    \small
\begin{tabular}{c|c|cc|ccc}
\Xhline{1.0pt}
\multirow{2}{*}{Dataset} & \multirow{2}{*}{Stage} & \multirow{2}{*}{AP$^\text{bbox}$} & \multirow{2}{*}{AP$^\text{mask}$}  & \multicolumn{3}{c}{cosine distance} \\
\cline{5-7}
 &  &  &  & input & output  & att \\
\hline
\multirow{3}{35pt}{\centering COCO}
& c3 & 37.5 & 34.4 & 0.326 & 0.004 & 0.009 \\
& c4 & 38.1 & 34.8 & 0.401 & 0.012 & 0.020 \\
& c5 & 38.2 & 35.1 & 0.372 & 0.024 & 0.042 \\
 \hline
 & Stage & Top-1 & Top-5   & input & output  & att \\
\cline{2-7}
\multirow{2}{35pt}{\centering Kinetics}
& c3 & 75.5 & 92.1 & 0.297 & 0.007 & 0.005 \\
& c4 & 75.4 & 92.1 & 0.395 & 0.001 & 0.001 \\
 \hline
\multirow{3}{35pt}{\centering ImageNet}
& c3 & 77.0 & 93.4 & 0.248 & 0.067 & 0.041 \\
& c4 & 77.2 & 93.5 & 0.301 & 0.074 & 0.115 \\
& c5 & 76.5 & 93.2 & 0.257 & 0.013 & 0.033 \\
\Xhline{1.0pt}
\end{tabular}
    \caption{Statistical analysis of non-local block (Embedded Gaussian) at \textbf{different stages} on four tasks.}
\label{table:statistical-nlocal-stages}
\end{minipage}\hfill
\begin{minipage}{0.52\linewidth}
\centering
    \addtolength{\tabcolsep}{-1.5pt}
    \small
\begin{tabular}{c|c|c|cc|cc|c}
\Xhline{1.0pt}
\multirow{2}{35pt}{\centering Dataset}& \multirow{2}{35pt}{\centering Method} & \multicolumn{6}{c}{cosine distance} \\
\cline{3-8}
 &  & input & {key} & {query} & prod & att & output  \\
\hline
\multirow{2}{35pt}{\centering COCO}
& E-Gaussian & 0.401 & 0.332 & 0.050 & 0.005 & 0.020 & 0.012 \\
& Gaussian & 0.397 & - & - & 0.069 & 0.177 & 0.062 \\
\hline
\multirow{2}{35pt}{\centering Kinetics}
& E-Gaussian & 0.358 & 0.356 & 0.264 & 0.404 & 0.004 & 0.003 \\
& Gaussian & 0.345 & - & - & 0.036 & 0.056 & 0.056 \\
\hline
\multirow{2}{35pt}{\centering ImageNet}
& E-Gaussian & 0.301 & 0.234 & 0.156 & 0.340 & 0.115 & 0.074 \\
& Gaussian & 0.045 & - & - & 0.001 & 0.011 & 0.005 \\
\Xhline{1.0pt}
\end{tabular}
\caption{\textbf{Fine-grained statistical analysis} of non-local block (Embedded Gaussian and Gaussian) on four tasks.}
\label{table:statistical-nlocal-finegrained}
\end{minipage}
\vspace{-12pt}
\end{table*}

\subsubsection{Statistical Analysis}
To more rigorously verify the phenomenon observed from the visualization, we statistically compare the differences (cosine distances) between the input features and the output features of different positions. Denote ${\bf v}_i$ as the feature vector for position $i$. The average distance measure is defined as 
$avg\_dist = \frac{1}{{N_p^2}} \sum\nolimits_{i = 1}^{{N_p}} {\sum\nolimits_{j = 1}^{{N_p}} {dist\left( {{{\bf v} _i},{{\bf v} _j}} \right)} }$, 
where $dist(\cdot,\cdot)$ is the distance function between two vectors. 
\textbf{Cosine distance} is a widely-used distance measure, defined as $dist({{\bf v} _i},{{\bf v} _j})$=$(1-\cos({{\bf v} _i},{{\bf v} _j}))/2$.

\noindent \textbf{Different Non-local Instantiations/Tasks.} The average cosine distances are computed between input features, attention maps and output features of different positions, with four instantiations of the non-local block on three standard tasks: object detection on COCO, action recognition on Kinetics, and image classification on ImageNet.
In detail, we compute the cosine distance between three kinds of vectors, the non-local block inputs (${\bf v}_i\leftarrow{\bf x}_i$, `input' in Table \ref{table:statistical-nlocal}), the non-local block outputs before fusion (${\bf v}_i\leftarrow{\bf z}_i-{\bf x}_i$, `output' in Table \ref{table:statistical-nlocal}), and the attention maps of query positions (${\bf v}_i\leftarrow{\omega}_i$, `att' in Table \ref{table:statistical-nlocal}).

Results with four instantiations of the non-local block on four standard tasks are shown in Table \ref{table:statistical-nlocal}.
First, large values of cosine distance in the `input' column show that the input features for the non-local block are discriminative across different positions. But the values of cosine distance in the `output' column are at least one order of magnitude smaller than that in the 'input' column on COCO, Kinetics and ImageNet, indicating that output global context features modeled by the non-local block on these three tasks are almost the same for different query positions.
The cosine distances on attention maps (`att') are also very small for all instantiations on these three tasks, which again verifies the observation from the visualization.

To conclude, although a non-local block intends to compute the global context specific to each query position, the global context after training is actually independent of query position.
Hence, it may be redundant for the non-local block to compute different attention maps for different positions, allowing us to simplify the non-local block.

\noindent \textbf{Insertion at Different Stages.} 
It is widely accepted that the lower layers of deep networks contain low-level, less-semantic features such as local edges, and higher layers contain high-level features with more semantic information, such as parts and objects \cite{DBLP:journals/corr/ZeilerF13}.
The non-local block may perform differently at different places in a deep network.
To examine this, we have also done a statistical analysis across different stages with the most widely-used instantiation, Embedded Gaussian, on the four standard tasks.

For different tasks, the non-local block is applied at different positions. For example, in action recognition on kinetics, the non-local blocks are inserted only in c4 and c5, hence we perform the experiments accordingly.

Results are presented in Table~\ref{table:statistical-nlocal-stages}.
Interestingly, we can see an obvious trend from lower layers to higher layers, that the output features in higher layers are more query-dependent than that in the lower layers.

\noindent \textbf{Fine-grained Analysis.} To analyze the reason of this phenomenon, we have done a more fine-grained statistical analysis on the two most widely-adopted instantiations of the non-local block, Embedded Gaussian and Gaussian. 

For Embedded Gaussian, we compute the average cosine distances between input features (input), features after the $W_k$ transform (key), features after the $W_q$ transform (query), different query features after inner product (prod), attention maps (att), and output features (output), which are marked in Figure~\ref{fig:arch-nl} (a).
For Gaussian, as marked in Figure~\ref{fig:arch-nl} (b), we compute the average cosine distances between input features (input), different query features after inner product (prod), attention maps (att), and output features (output).

Results of the fine-grained statistical analysis are shown in Table~\ref{table:statistical-nlocal-finegrained}.
First, we look into the results on COCO, Kinetics and ImageNet.
For Embedded Gaussian, although $W_q$ and $W_k$ are both 1x1 convolutions with the same input, the features after $W_q$ are more similar, and the features after $W_k$ are still different.
Also, features after the inner-product computation are more query-independent after training.
For Gaussian, as this instantiation does not include the query and key transformations, the attention maps still appear query-dependent. But after attention pooling and the output transform, the differences between the output features are significantly reduced, and are almost one order of magnitude smaller than that of the input features.

In our understanding, the tasks drive the network components to learn the specific architecture that can benefit the tasks most.
And query-independence of the non-local block can benefit three major tasks: object detection on COCO, action recognition on Kinetics, and image recognition on ImageNet.

\begin{table}[]
    \centering
    \addtolength{\tabcolsep}{-3pt}
    \small
\begin{tabular}{c|cc|ccc}
\Xhline{1.0pt}
 Method & mIoU  &   & input & output  & att \\
 \hline
 Gaussian & 76.47 &  & 0.318 & 0.433 & 0.478 \\
 E-Gaussian & 77.59 &  & 0.315 & 0.393 & 0.354 \\
 Dot product & 77.74 &  & 0.323 & 0.386 & 0.331 \\
  Concat & 77.78 &  & 0.321 & 0.002 & 0.001 \\
\Xhline{1.0pt}
\end{tabular}
    \caption{Statistical analysis using four instantiations of non-local blocks on Cityscape semantic segmentation. `input' denotes the input of the non-local block (${\bf x}_i$), `output' denotes the output of the non-local block (${\bf z}_i-{\bf x}_i$), `att' denotes the attention map of query positions ($\omega_i$).}
\label{table:statistical-nlocal-seg}
\end{table}

\noindent \textbf{Exceptions.} Although non-local networks do not learn pairwise relations on the above three important visual recognition tasks, we note that there are also some tasks where non-local networks successfully learn pairwise relations, e.g. semantic segmentation on Cityscapes, as illustrated in Table~\ref{table:statistical-nlocal-seg}. Table~\ref{table:ablation-cityscapes} also shows that NLNet can improve segmentation accuracy over the regular counterpart. A question is whether such improvements are due mainly to the learnt pairwise relations. Surprisingly, a simplified version of NLNet (noted as SNL, which will be introduced in the next section) which models only global context also shows performance comparable to NLNet. This indicates that although the non-local block applied in semantic segmentation may learn pairwise relations, the accuracy improvement may be mostly ascribed to the modeling of global context.

\begin{figure*}[!htb]
    \centering
    \includegraphics[width=0.85\linewidth]{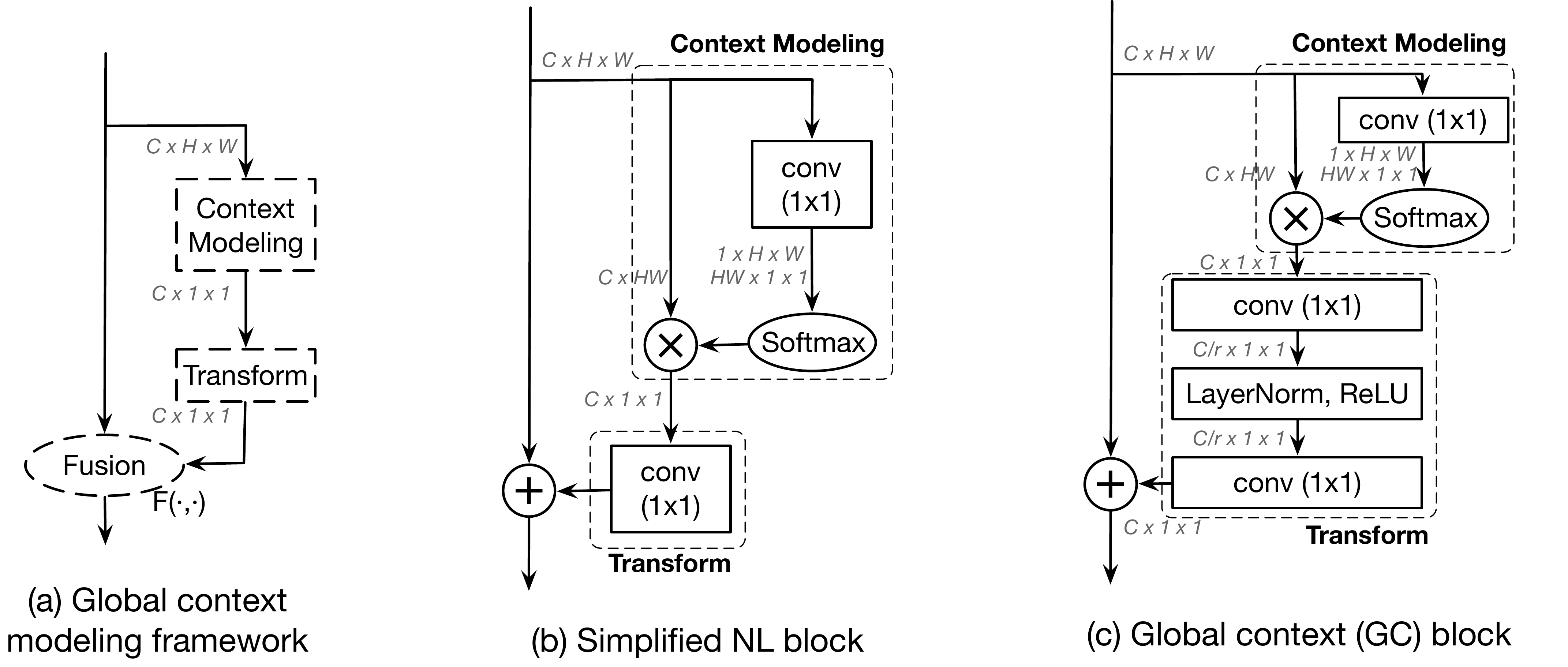}
	\caption{\textbf{Architecture of the main blocks}. The feature maps are shown as feature dimensions, e.g. CxHxW denotes a feature map with channel number C, height H and width W. $\otimes$ denotes matrix multiplication, $\oplus$ denotes broadcast element-wise addition, and $\odot$ denotes broadcast element-wise multiplication.}
	\label{fig:arch}
\end{figure*}

\section{Method}

\jiarui{In the last section, both qualitative and statistical analysis indicate that non-local blocks tend to learn query-independent attention maps in many visual recognition tasks, instead of query-dependent context as implied by the formulation. This finding challenges the necessity of the query-dependent formulation in the original non-local block, and raises the question of whether explicit query-independent attention maps perform worse than the original query-dependent formulation. We answer this in the following subsections. We first present a simplified non-local formulation by explicitly making the attention maps query-independent in Section~\ref{sec.snl}. We will show in experiments that this simplified formulation can significantly reduce computation yet maintain accuracy. Then in Section~\ref{sec.framework}, we abstract this simplified non-local formulation into a general global context modeling framework, which interestingly also operates like the popular SE block~\cite{hu2018senet}. Finally, in Section~\ref{sec.gcblock}, we present our global context block, which is a new instantiation of the general framework by combining the strengths of the simplified non-local block and the SE block~\cite{hu2018senet}.}

\subsection{Simplifying the Non-local Block}
\label{sec.snl}
As the widely-adopted Embedded Gaussian instantiation achieves representative 
performance on all three standard tasks, as shown in Table \ref{table:statistical-nlocal}, we adopt the Embedded Gaussian as the basic non-local block in the following sections.
Based on the observation that the attention maps for different query positions are almost the same, we simplify the non-local block by computing a global (query-independent) attention map and share this global attention map among all query positions. 
Following the results in \cite{hu2018relation} that variants with and without $W_z$ achieve comparable performance, we omit $W_z$ in the simplified version.
Our simplified non-local block is defined as
\begin{equation}\label{eqn:snlocal}
    {{\bf{z}}_i} = {{\bf{x}}_i} + \sum\nolimits_{j=1}^{N_p} {\frac{{\exp \left( {{W_k}{{\bf{x}}_j}} \right)}}{{\sum\nolimits_{m=1}^{N_p} {\exp \left( {{W_k}{{\bf{x}}_m}} \right)} }}\left( {{W_v} \cdot {{\bf{x}}_j}} \right)},
\end{equation}
where $W_k$ and $W_v$ denote linear transformation matrices. 

To further reduce the computational cost of this simplified block, we apply the distributive law to move $W_v$ outside of the attention pooling, as
\begin{equation}\label{eqn:snlocal2}
    {{\bf{z}}_i} = {{\bf{x}}_i} + {W_v}\sum\nolimits_{j=1}^{N_p} {\frac{{\exp \left( {{W_k}{{\bf{x}}_j}} \right)}}{{\sum\nolimits_{m=1}^{N_p} {\exp \left( {{W_k}{{\bf{x}}_m}} \right)} }}{{\bf{x}}_j}}.
\end{equation}
This version of simplified non-local block is illustrated in Figure \ref{fig:arch}(b). 
After moving $W_v$ outside of attention pooling, the FLOPs of this 1x1 convolution $W_v$ is reduced from $\mathcal{O}$(HWC$^2$) to $\mathcal{O}$(C$^2$).

Different from the traditional non-local block, the second term in Eqn. \ref{eqn:snlocal2} is independent of the query position $i$, which means that this term is shared across all query positions $i$. 
We thus directly model global context as a weighted sum of the features at all positions, and aggregate (add) the global context features to the features at each query position.
In experiments, we directly replace the non-local (NL) block with our simplified non-local (SNL) block, and evaluate accuracy and computation cost on four tasks, object detection on COCO, semantic segmentation on Cityscapes, ImageNet classification, and action recognition on Kinetics, shown in Tables \ref{table:ablation-coco}(a), \ref{table:ablation-imagenet}(a), \ref{table:ablation-cityscapes}(a) and \ref{table:ablation-kinetics}.
As expected, the SNL block achieves performance comparable to (or slightly below) the NL block with significantly lower FLOPs.

\subsection{Global Context Modeling Framework}
\label{sec.framework}
As shown in Fig.~\ref{fig:arch}(b), the simplified non-local block can be abstracted into three parts: (a) global attention pooling, which adopts a 1x1 convolution $W_k$ and a softmax function to obtain the attention weights, and then performs attention pooling to obtain the global context features; (b) feature transform via a 1x1 convolution $W_v$; (c) feature aggregation, which employs addition to aggregate global context features to each position.

We regard this abstraction as a global context modeling framework, illustrated in Figure \ref{fig:arch}(a) and defined as
\begin{equation}\label{eqn:framework}
    {{\bf{z}}_i} = F\left( {{{\bf{x}}_i},\delta \left( {\sum\nolimits_{j=1}^{N_p} {{\alpha _j}{{\bf{x}}_j}} } \right)} \right),
\end{equation}
where (a) ${\sum\nolimits_j {{\alpha _j}{{\bf{x}}_j}} }$ denotes the \textbf{context modeling} module which groups the features of all positions together via weighted averaging with weight $\alpha_j$ to obtain the global context features (global attention pooling in the simplified NL (SNL) block);
(b) $\delta(\cdot)$ denotes the feature \textbf{transform} to capture channel-wise dependencies (1x1 convolution in the SNL block); 
and (c) $F(\cdot,\cdot)$ denotes the \textbf{fusion} function to aggregate the global context features to the features of each position (broadcast element-wise addition in the SNL block).

Interestingly, the squeeze-excitation (SE) block proposed in \cite{hu2018senet} is also an instantiation of our proposed framework, which consists of: 
(a) global average pooling for global context modeling (set $\alpha_j=\frac{1}{N_p}$ in Eqn. \ref{eqn:framework}), called the squeeze operation in the SE block; 
(b) a bottleneck transform module (let $\delta(\cdot)$ in Eqn. \ref{eqn:framework} be one 1x1 convolution, one ReLU, one 1x1 convolution and a sigmoid function, sequentially), to compute the importance for each channel, called the excitation operation in the SE block; and 
(c) a rescaling function for fusion (let $F(\cdot,\cdot)$ in Eqn. \ref{eqn:framework} be element-wise multiplication), to recalibrate the channel-wise features.

\subsection{Global Context Block}
\label{sec.gcblock}
Here we propose a new instantiation of the global context modeling framework, named the global context (GC) block, which can effectively model long-range dependency as a simplified non-local block, and is lightweight for application to all layers with a small increase in FLOPs.

In the simplified non-local block, shown in Figure \ref{fig:arch}(b), the transform module has the largest number of parameters, including from one 1x1 convolution with C$\cdot$C parameters.
When we add this SNL block to higher layers, e.g. res$_5$, the number of parameters of this 1x1 convolution, C$\cdot$C=2048$\cdot$2048, dominates the number of parameters of this block.
Hence, this 1x1 convolution is replaced by a bottleneck transform module, which significantly reduces the number of parameters from C$\cdot$C to 2$\cdot$C$\cdot$C/r, where r is the bottleneck ratio and C/r denotes the hidden representation dimension of the bottleneck.
With the default reduction ratio set to r=16, the number of parameters for the transform module can be reduced to 1/8 of the original SNL block. More results on different values of bottleneck ratio r are shown in Table \ref{table:ablation-coco}(e).

As the two-layer bottleneck transformation increases the difficulty of optimization, we add layer normalization inside the bottleneck transformation (before ReLU) to ease optimization, as well as to act as a regularizer that can benefit generalization.
As shown in Table \ref{table:ablation-coco}(d), layer normalization can significantly enhance the performance of object detection and segmentation on COCO.

The detailed architecture of the global context (GC) block is illustrated in Figure \ref{fig:arch}(c) and formulated as
\begin{equation}
\small
{{\bf{z}}_i} = {{\bf{x}}_i} + {W_{v2}}{\rm{ReLU}}\bigg( {{\rm{LN}}\bigg( {{W_{v1}}\sum\limits_{j=1}^{N_p} {\frac{{{e^{{W_k}{{\bf{x}}_j}}}}}{{\sum\limits_{m=1}^{N_p} {{e^{{W_k}{{\bf{x}}_m}}}} }}{{\bf{x}}_j}} } \bigg)} \bigg),
\end{equation}
where $\alpha_j=\frac{{{e^{{W_k}{{\bf{x}}_j}}}}}{{\sum\nolimits_m {{e^{{W_k}{{\bf{x}}_m}}}} }}$ is the weight for global attention pooling, and $\delta(\cdot)=W_{v2}{\rm{ReLU}}({\rm{LN}}(W_{v1}(\cdot)))$ denotes the bottleneck transform.
Specifically, our GC block consists of: (a) global attention pooling for context modeling; (b) bottleneck transform to capture channel-wise dependencies; and (c) broadcast element-wise addition for feature fusion.

Since the GC block is lightweight, it can be applied in multiple layers to better capture long-range dependency with only a slight increase in computation cost. Taking ResNet-50 for ImageNet classification as an example, GC-ResNet-50 denotes adding the GC block to all layers (c3+c4+c5) in ResNet-50 with a bottleneck ratio of 16. GC-ResNet-50 increases ResNet-50 computation from $\sim$3.86 GFLOPs to $\sim$3.87 GFLOPs, corresponding to a 0.26\% relative increase.
Also, GC-ResNet-50 introduces $\sim$2.52M additional parameters beyond the $\sim$25.56M parameters required by ResNet-50, corresponding to a $\sim$9.86\% increase.

Global context can benefit a wide range of visual recognition tasks, and the flexibility of the GC block allows it to be plugged into network architectures used in various computer vision problems. In this paper, we apply our GC block to four general vision tasks -- image recognition, object detection/instance segmentation, semantic segmentation and action recognition -- and observe significant improvements in all four.

\noindent \textbf{Relationship to non-local block.}
As the non-local block actually learns query-independent global context, the global attention pooling of our global context block models the same global context as the NL block but with significantly lower computation cost.
As the GC block adopts the bottleneck transform to reduce redundancy in the global context features, the number of parameters and FLOPs are further reduced.
The FLOPs and number of parameters of the GC block are significantly lower than that of the NL block, allowing our GC block to be applied to multiple layers with just a slight increase in computation, while better capturing long-range dependency and aiding network training.

\noindent \textbf{Relationship to squeeze-excitation block.}
The main difference between the SE block and our GC block is the fusion module, which reflects the different goals of the two blocks. 
The SE block adopts rescaling to recalibrate the importance of channels but inadequately models long-range dependency.
Our GC block follows the NL block by utilizing addition to aggregate global context to all positions for capturing long-range dependency.
A second difference is with the layer normalization in the bottleneck transform.
As our GC block adopts addition for fusion, layer normalization can ease optimization of the two-layer architecture for the bottleneck transform, which can lead to better performance. 
Third, global average pooling in the SE block is a special case of global attention pooling in the GC block. 
Results in Tables \ref{table:ablation-coco}(d), \ref{table:ablation-coco}(f) and \ref{table:ablation-imagenet}(b) show the superiority of addition in the fusion module, layer normalization in the two-layer bottleneck, and the global attention pooling, compared to the SE block, respectively.

\section{Experiments}
To evaluate the proposed method, we carry out experiments on four basic tasks: object detection/instance segmentation on COCO \cite{lin2014coco}, image classification on ImageNet \cite{deng2009imagenet}, action recognition on Kinetics \cite{kay2017kinetics}, and semantic segmentation on Cityscapes \cite{cordts2016cityscapes}.
Experimental results demonstrate that the proposed GCNet generally outperforms non-local networks with significantly lower FLOPs.

\subsection{Object Detection/Instance Segmentation on COCO}
We investigate our model on object detection and instance segmentation on COCO 2017 \cite{lin2014coco}, whose train set is comprised of 118k images, validation set of 5k images, and test-dev set of 20k images.
We follow the standard setting \cite{he2017mask} of evaluating object detection and instance segmentation via the standard mean average-precision scores at different
boxes and the mask IoUs.

\textbf{Setup.}
Our experiments are implemented with PyTorch \cite{paszke2017pytorch} based on open source mmdetection \cite{chen2019mmdetection}.
Unless otherwise noted, our GC block of ratio $r$=16 is applied to stages c3, c4, c5 of ResNet/ResNeXt.

\textbf{Training.}
We use the standard configuration of Mask R-CNN \cite{he2017mask} with FPN and ResNet/ResNeXt as the backbone architecture.
The input images are resized such that their shorter side is of 800 pixels \cite{he2017fpn}.
We trained on 8 GPUs with 2 images per GPU (effective mini batch size of 16).
The backbones of all models are pretrained on ImageNet classification \cite{deng2009imagenet}, then all layers except for c1 and c2 are jointly finetuned with detection and segmentation heads.
Unlike stage-wise training with respect to RPN in \cite{he2017mask}, end-to-end training like in \cite{ren2015faster} is adopted for our implementation, yielding better results.
Different from the conventional finetuning setting \cite{he2017mask}, we use Synchronized BatchNorm to replace frozen BatchNorm.
All models are trained for 12 epochs using Synchronized SGD with a weight decay of 0.0001 and momentum of 0.9, which roughly corresponds to the 1x schedule in the Mask R-CNN benchmark \cite{massa2018mrcnn}.
The learning rate is initialized to 0.02, and decays by a factor of 10 at the 8th and 11th epochs.
The choice of hyper-parameters also follows the latest release of the Mask R-CNN benchmark~\cite{massa2018mrcnn}.

\begin{table}[]
    \centering
    \footnotesize
    \addtolength{\tabcolsep}{-5.5pt}
\begin{tabular}{c|ccc|ccc|c|c}
\Xhline{1.0pt}
\multicolumn{9}{c}{(a) \textbf{Block design}}                              \\
     & AP${^\text{bbox}}$ & AP$^\text{bbox}_\text{50}$ & AP$^\text{bbox}_{75}$&AP$^\text{mask}$&AP$^\text{mask}_\text{50}$&AP$^\text{mask}_\text{75}$ &  \#param & FLOPs \\
\hline
    baseline & 37.2   & 59.0 & 40.1 & 33.8 & 55.4 & 35.9 & 44.4M & 279.4G \\
+1 NL & 38.0 & 59.8 & 41.0 & 34.7 & 56.7 & 36.6 & 46.5M & 288.7G \\
+1 SNL & 38.1 & 60.0 & 41.6 & 35.0 & 56.9 & 37.0 & 45.4M & 279.4G \\
+1 GC & 38.1  & 60.0 & 41.2 & 34.9 & 56.5 & 37.2 & 44.5M & 279.4G \\
+all GC  & \textbf{39.4} & \textbf{61.6} & \textbf{42.4} & \textbf{35.7} & \textbf{58.4} & \textbf{37.6} & 46.9M & 279.6G \\
\Xhline{1.0pt}
\multicolumn{9}{c}{(b) \textbf{Positions}}                                     \\
     & AP${^\text{bbox}}$ & AP$^\text{bbox}_\text{50}$ & AP$^\text{bbox}_\text{75}$&AP$^\text{mask}$&AP$^\text{mask}_\text{50}$&AP$^\text{mask}_\text{75}$ &  \#param & FLOPs \\
\hline
    baseline & 37.2   & 59.0 & 40.1 & 33.8 & 55.4 & 35.9 & 44.4M & 279.4G \\
afterAdd & \textbf{39.4} & \textbf{61.9} & \textbf{42.5} & \textbf{35.8} & \textbf{58.6} & \textbf{38.1} & 46.9M & 279.6G \\
after1x1 & \textbf{39.4} & 61.6 & 42.4 & 35.7 & 58.4 & 37.6 & 46.9M & 279.6G \\
\Xhline{1.0pt}
\multicolumn{9}{c}{(c) \textbf{Stages}}                                       \\
     & AP${^\text{bbox}}$ & AP$^\text{bbox}_\text{50}$ & AP$^\text{bbox}_\text{75}$&AP$^\text{mask}$&AP$^\text{mask}_\text{50}$&AP$^\text{mask}_\text{75}$ &  \#param & FLOPs \\
\hline
    baseline & 37.2   & 59.0 & 40.1 & 33.8 & 55.4 & 35.9 & 44.4M & 279.4G \\
c3 & 37.9 & 59.6 & 41.1 & 34.5 & 56.3 & 36.8 & 44.5M & 279.5G \\
c4 & 38.9 & 60.9 & 42.2 & 35.5 & 57.6 & 37.7 & 45.2M & 279.5G \\
c5 & 38.7 & 61.1 & 41.7 & 35.2 & 57.4 & 37.4 & 45.9M & 279.4G \\
c3+c4+c5 & \textbf{39.4} & \textbf{61.6} & \textbf{42.4} & \textbf{35.7} & \textbf{58.4} & \textbf{37.6} & 46.9M & 279.6G \\
\Xhline{1.0pt}
\multicolumn{9}{c}{(d) \textbf{Bottleneck design}}                            \\
     & AP${^\text{bbox}}$ & AP$^\text{bbox}_\text{50}$ & AP$^\text{bbox}_\text{75}$&AP$^\text{mask}$&AP$^\text{mask}_\text{50}$&AP$^\text{mask}_\text{75}$ &  \#param & FLOPs \\
\hline
    baseline & 37.2   & 59.0 & 40.1 & 33.8 & 55.4 & 35.9 & 44.4M & 279.4G \\
w/o ratio & \textbf{39.4} & \textbf{61.8} & \textbf{42.8} & \textbf{35.9} & \textbf{58.6} & \textbf{38.1} & 64.4M & 279.6G \\
r16 (ratio 16) & 38.8 & 61.0 & 42.3 & 35.3 & 57.6 & 37.5 & 46.9M & 279.6G \\
r16+ReLU & 38.8 & 61.0 & 42.0 & 35.4 & 57.5 & 37.6 & 46.9M & 279.6G \\
r16+LN+ReLU & \textbf{39.4} & 61.6 & 42.4 & 35.7 & 58.4 & 37.6 & 46.9M & 279.6G \\
\Xhline{1.0pt}
\multicolumn{9}{c}{(e) \textbf{Bottleneck ratio}}                                        \\
     & AP${^\text{bbox}}$ & AP$^\text{bbox}_\text{50}$ & AP$^\text{bbox}_\text{75}$&AP$^\text{mask}$&AP$^\text{mask}_\text{50}$&AP$^\text{mask}_\text{75}$ &  \#param & FLOPs \\
\hline
    baseline & 37.2   & 59.0 & 40.1 & 33.8 & 55.4 & 35.9 & 44.4M & 279.4G \\
ratio 4  & \textbf{39.9} & \textbf{62.2} & \textbf{42.9} & \textbf{36.2} & \textbf{58.7} & \textbf{38.3} & 54.4M & 279.6G \\
ratio 8  & 39.5 & 62.1  & 42.5 & 35.9 & 58.1 & 38.1 & 49.4M & 279.6G \\
ratio 16 & 39.4 & 61.6 & 42.4 & 35.7 & 58.4 & 37.6 & 46.9M & 279.6G \\
ratio 32 & 39.1 & 61.6 & 42.4 & 35.7 & 58.1 & 37.8 & 45.7M & 279.5G \\
\Xhline{1.0pt}
\multicolumn{9}{c}{(f) \textbf{Pooling and fusion}}                                \\
     & AP${^\text{bbox}}$ & AP$^\text{bbox}_\text{50}$ & AP$^\text{bbox}_\text{75}$&AP$^\text{mask}$&AP$^\text{mask}_\text{50}$&AP$^\text{mask}_\text{75}$ &  \#param & FLOPs \\
\hline
    baseline & 37.2   & 59.0 & 40.1 & 33.8 & 55.4 & 35.9 & 44.4M & 279.4G \\
avg+scale & 38.2 & 60.2 & 41.2 & 34.7 & 56.7 & 37.1 & 46.9M & 279.5G \\
avg+add & 39.1 & 61.4 & 42.3 & 35.6 & 57.9 & \textbf{37.9} & 46.9M & 279.5G \\
att+scale & 38.3 & 60.4 & 41.5 & 34.8 & 57.0 & 36.8 & 46.9M & 279.6G \\
att+add & \textbf{39.4} & \textbf{61.6} & \textbf{42.4} & \textbf{35.7} & \textbf{58.4} & 37.6 & 46.9M & 279.6G \\
\Xhline{1.0pt}
\end{tabular}
\caption{\textbf{Ablation study} based on Mask R-CNN, using ResNet-50 as backbone with FPN, for \textbf{object detection} and \textbf{instance segmentation} on COCO 2017 validation set.}
	\label{table:ablation-coco}
\end{table}

\begin{table}[t]
\footnotesize
\centering
\addtolength{\tabcolsep}{-5pt}
\begin{tabular}[t]{ccc|ccc|ccc}
\Xhline{1.0pt}
\multicolumn{9}{c}{(a) \textbf{Different Normalization}}                              \\
     backbone & head & method &
     AP$^\text{bbox}$ & AP$^\text{bbox}_\text{50}$ & AP$^\text{bbox}_\text{75}$ &
     AP$^\text{mask}$ & AP$^\text{mask}_\text{50}$ & AP$^\text{mask}_\text{75}$ \\
    \hline
    \multirow{3}{*}{fixBN} & \multirow{2}{*}{2fc} & baseline & 37.3 & 59.0 & 40.2 & 34.2 & 55.9 & 36.2 \\ 
                           & \multirow{2}{*}{(w/o BN)} & +GC r16 & 38.5 & 60.8 & 41.5 & 35.1 & 57.3 & 37.1 \\
                           &  & +GC r4& \textbf{38.9} & \textbf{61.1} & \textbf{42.0} & \textbf{35.5} & \textbf{57.7} & \textbf{37.5} \\
    \hline
    \multirow{3}{*}{syncBN} & \multirow{2}{*}{2fc} & baseline & 37.2 & 59.0 & 40.1 & 33.8 & 55.4 & 35.9 \\
                            & \multirow{2}{*}{(w/o BN)} & +GC r16 & 39.4 & 61.6 & 42.4 & 35.7 & 58.4 & 37.6 \\
                            & & +GC r4& \textbf{39.9} & \textbf{62.2} & \textbf{42.9} & \textbf{36.2} & \textbf{58.7} & \textbf{38.3} \\
    \hline
    \multirow{3}{*}{syncBN} & \multirow{2}{*}{4conv1fc} & baseline & 38.8 & 59.5 & 42.6 & 34.6 & 56.2 & 37.1 \\
                           & \multirow{2}{*}{syncBN} & +GC r16 & 41.0 & 62.1 & 44.9 & 36.5 & 58.3 & 39.0\\ 
                           & & +GC r4 & \textbf{41.4} & \textbf{62.5} & \textbf{45.5} & \textbf{37.0} & \textbf{59.1} & \textbf{39.5}\\
    \Xhline{1.0pt}
    \multicolumn{9}{c}{(b) \textbf{Longer Training}} \\
    setting & schd & method & 
     AP$^\text{bbox}$ & AP$^\text{bbox}_\text{50}$ & AP$^\text{bbox}_\text{75}$ &
     AP$^\text{mask}$ & AP$^\text{mask}_\text{50}$ & AP$^\text{mask}_\text{75}$ \\
    \hline
    \multirow{2}{*}{syncBN} & \multirow{3}{*}{2x} & baseline & 37.7 & 59.1 & 40.9 & 34.3 & 55.8 & 36.5 \\
    \multirow{2}{*}{2fc} &  & +GC r16 & 39.7 & 61.8 & 43.0 & 36.0 & 58.5 & 38.4 \\
     &  & +GC r4 & \textbf{40.2} & \textbf{62.2} & \textbf{43.5} & \textbf{36.3} & \textbf{58.6} & \textbf{38.5}\\
    \Xhline{1.0pt}
    \end{tabular}
    \caption{\textbf{Ablation study} on different normalization and training schedules for \textbf{object detection} and \textbf{instance segmentation} on COCO 2017 validation set.}
	\label{table:bn-coco}
  \end{table}

\subsubsection{Ablation Study}
The ablation study is done on the COCO 2017 validation set. The standard COCO metrics including AP, AP$_{\text{50}}$, AP$_{\text{75}}$ for both bounding boxes and segmentation masks are reported.

\textbf{Block design.}
Following \cite{wang2017non}, we insert 1 non-local block (NL), 1 simplified non-local block (SNL), or 1 global context block (GC) right before the last residual block of c4.
Table \ref{table:ablation-coco}(a) shows that both SNL and GC achieve performance comparable to NL with fewer parameters and less computation,
indicating redundancy in computation and parameters in the original non-local design.
Furthermore, adding the GC block in all residual blocks yields higher performance (1.1\%$\uparrow$ on AP$^\text{bbox}$ and 0.9\%$\uparrow$ on AP$^\text{mask}$) with a slight increase in FLOPs and \#params.

\textbf{Positions.}
The NL block is inserted after the residual block (afterAdd), while the SE block is integrated after the last 1x1 convolution inside the residual block (after1x1).
In Table \ref{table:ablation-coco}(b), we investigate both cases with the GC block and they yield similar results. Hence, we adopt after1x1 as the default.

\textbf{Stages.}
Table \ref{table:ablation-coco}(c) shows the results of integrating the GC block at different stages.
All stages benefit from global context modeling in the GC block (0.7\%-1.7\%$\uparrow$ on AP$^\text{bbox}$ and AP$^\text{mask}$).
Inserting into c4 and c5 both achieves better performance than into c3, demonstrating that better semantic features can benefit more from the global context modeling.
With a slight increase in FLOPs, inserting the GC block into all layers (c3+c4+c5) yields even higher performance than inserting into only a single layer.

\textbf{Bottleneck design.}
The effects of each component in the bottleneck transform are shown in Table \ref{table:ablation-coco}(d).
w/o ratio denotes the simplified NLNet using one 1x1 convolution as the transform, which has more parameters compared to the baseline.
Even though r16 and r16+ReLU have much fewer parameters than the w/o ratio variant, two layers are found to be harder to optimize and lead to worse performance than a single layer.
So LayerNorm (LN) is exploited to ease optimization, leading to performance similar to w/o ratio but with much fewer \#params. 

\jiarui{The reason we adopt layer norm here is that other alternatives, i.e. batch norm and group norm, do not perform well probably due to insufficient statistics to compute the means and variances. The spatial resolution of the intermediate feature map in the GC block has been reduced to $1\times 1$ (see Fig 4(c)). If batch normalization is used, the number of elements to compute each mean and variance is $b$ ($b$ is the batch size), which is small. If group normalization is used, the number of elements to compute each mean and variance is $C/r/g$ ($g$ is the group number), which is also small. 
For layer norm, the number of elements used to compute each mean and variance is $C/r$, which is observed to be sufficient.}

\textbf{Bottleneck ratio.}
The bottleneck design is intended to reduce redundancy in parameters and provide a good tradeoff between performance and \#params.
In Table \ref{table:ablation-coco}(e), we vary the ratio r of the bottleneck.
As the ratio r decreases (from 32 to 4) with increasing number of parameters and FLOPs, the performance improves consistently (0.8\%$\uparrow$ on AP$^\text{bbox}$ and 0.5\%$\uparrow$ on AP$^\text{mask}$), indicating that our bottleneck strikes a good balance between performance and number of parameters.
It is worth noting that even with a ratio of r=32, the network still outperforms baseline by large margins.

\begin{table}[t]
    \footnotesize
    \centering
    \addtolength{\tabcolsep}{-6.2pt}
    \begin{tabular}[t]{cc|ccc|ccc|cc}
    \Xhline{1.0pt}
\multicolumn{9}{c}{(a) \textbf{test on validation set}}                            \\
     \multicolumn{2}{c|}{\centering backbone} &
     AP$^\text{bbox}$ & AP$^\text{bbox}_\text{50}$ & AP$^\text{bbox}_\text{75}$ &
     AP$^\text{mask}$ & AP$^\text{mask}_\text{50}$ & AP$^\text{mask}_\text{75}$ &
     FLOPS \\
    \hline
    \multirow{3}{*}{R50}              & baseline & 37.2 & 59.0 & 40.1 & 33.8 & 55.4 & 35.9 & 279.4G  \\
                                      & +GC r16 & 39.4 & 61.6 & 42.4 & 35.7 & 58.4 & 37.6 &  279.6G \\
                                      & +GC r4& \textbf{39.9} & \textbf{62.2} & \textbf{42.9} & \textbf{36.2} & \textbf{58.7} & \textbf{38.3} & 279.6G \\
    \hline
    \multirow{3}{*}{R101}             & baseline & 39.8 & 61.3 & 42.9 & 36.0 & 57.9 & 38.3 & 354.0G \\
                                      & +GC r16 & 41.1 & 63.6 & 45.0 & 37.4 & 60.1 & 39.6 & 354.3G \\
                                      & +GC r4 & \textbf{41.7} & \textbf{63.7} & \textbf{45.5} & \textbf{37.6} & \textbf{60.5} & \textbf{39.8} & 354.3G \\
    \hline
    \multirow{3}{*}{X101}             & baseline & 41.2 & 63.0 & 45.1 & 37.3 & 59.7 & 39.9 & 357.9G \\
                                      & +GC r16 & 42.4 & 64.6 & 46.5 & 38.0 & 60.9 & 40.5 & 358.2G \\
                                      & +GC r4 & \textbf{42.9} & \textbf{65.2} & \textbf{47.0} & \textbf{38.5} & \textbf{61.8} & \textbf{40.9} & 358.2G \\
    \hline
    \multirow{2}{*}{X101}             & baseline & 44.7 & 63.0 & 48.5 & 38.3 & 59.9 & 41.3 & 536.9G \\
    \multirow{2}{*}{+Cascade}         & +GC r16 & 45.9 & 64.8 & 50.0 & 39.3 & 61.8 & 42.1 & 537.2G \\
                                      & +GC r4 & \textbf{46.5} & \textbf{65.4} & \textbf{50.7} & \textbf{39.7} & \textbf{62.5} & \textbf{42.7} & 537.3G \\

    \hline
    \multirow{2}{*}{X101+DCN}         & baseline & 47.1 & 66.1 & 51.3 & 40.4 & 63.1 & 43.7 & 547.5G \\
    \multirow{2}{*}{+Cascade}         & +GC r16 & \textbf{47.9} & \textbf{66.9} & \textbf{52.2} & \textbf{40.9} & 63.7 & \textbf{44.1} & 547.8G \\
                                      & +GC r4 & \textbf{47.9} & \textbf{66.9} & 51.9 & 40.8 & \textbf{64.0} & 44.0 & 547.8G \\
    \hline
    X101 64x4d+DCN         & +Cascade &  &  &  &  &  &  &  \\
    + 4conv1fc head          & +GC r4  & \textbf{51.8} & \textbf{70.4} & \textbf{56.1} & \textbf{44.7} & \textbf{67.9} & \textbf{48.4} & 1040.6G \\
    + multiscale     +                  & 3x schd &  &  &  &  &  &  &  \\
\hline
\multicolumn{9}{c}{(b) \textbf{test on test-dev set}}  \\
    \hline

    \multirow{2}{*}{X101}             & baseline & 45.0 & 63.7	& 49.1 & 38.7 & 60.8 & 41.8 & 536.9G \\
    \multirow{2}{*}{+Cascade}         & +GC r16 & 46.5 & 65.7 & 50.7 & 40.0 & 62.9 & 43.1 & 537.2G \\
                                      & +GC r4 & \textbf{46.6} & \textbf{65.9} & \textbf{50.7} & \textbf{40.1} & \textbf{62.9} & \textbf{43.3} & 537.3G \\
    \hline
    \multirow{2}{*}{X101+DCN}         & baseline & 47.7 & 66.7 & 52.0 & 41.0 & 63.9 & 44.3 & 547.5G \\
    \multirow{2}{*}{+Cascade}         & +GC r16 & 48.3 & 67.5 & \textbf{52.7} & \textbf{41.5} & \textbf{64.6} & \textbf{45.0} & 547.8G \\
                                      & +GC r4 & \textbf{48.4} & \textbf{67.6} & \textbf{52.7} & \textbf{41.5} & \textbf{64.6} & \textbf{45.0} & 547.8G \\
    \hline
    X101 64x4d+DCN         & +Cascade &  &  &  &  &  &  &  \\
    + 4conv1fc head          & +GC r4  & \textbf{52.3} & \textbf{70.9} & \textbf{56.9} & \textbf{45.4} & \textbf{68.9} & \textbf{49.6} & 1040.6G \\
    + multiscale     +                  & 3x schd &  &  &  &  &  &  &  \\
    \Xhline{1.0pt}
    \end{tabular}
    \caption{Results of GCNet (ratio 4 and 16) with \textbf{stronger backbones} on COCO 2017 validation and test-dev sets.}
	\label{table:archs-coco}
  \end{table}

\textbf{Pooling and fusion.}
The different choices for pooling and fusion are ablated in Table \ref{table:ablation-coco}(f).
First, it shows that addition is more effective than scaling in the fusion stage.
It is surprising that attention pooling only achieves slightly better results than vanilla average pooling.
This indicates that how global context is aggregated to query positions (choice of fusion module) is more important than how features from all positions are grouped together (choice in context modeling module).
\jiarui{It is worth noting that att+add significantly outperforms avg+scale, which denotes the approach of SENet with layer norm, because of the effective modeling of long-range dependency with attention pooling for context modeling, and the use of addition for feature aggregation.}

\textbf{Different Normalization}
The result of different normalization is presented in \ref{table:bn-coco}(a). 
GCNet improves the performance by $1.0\% \uparrow$ on AP$^\text{bbox}$ and $0.7\% \uparrow$ on AP$^\text{mask}$ by replacing fixBN with syncBN in the backbone, 
while baseline maintains similar performance.
Since the backbone is already pretrained on ImageNet while the inserted GC block is randomly initialized, the running statistics of the backbone features could help with the training of the GC block. 
Following \cite{wu2018group, he2018rethinking}, syncBN is further applied in both the backbone and heads. 
Even though the baseline improves by $1.6\% \uparrow$ in AP$^\text{bbox}$ and $0.8\% \uparrow$ in AP$^\text{mask}$, 
the gap between GC and the baseline is still preserved, 
which is $2.6\% \uparrow$ in AP$^\text{bbox}$ and $2.4\% \uparrow$ in AP$^\text{mask}$. 

\textbf{Longer Training}
We also trained our model for 24 epochs which is roughly the same as the 2x schedule in \cite{massa2018mrcnn}. 
As shown in \ref{table:bn-coco}(b), GCNet does not saturate and greater performance gain is observed, which indicates the large potential capacity of GCNet. 

\subsubsection{Experiments on Stronger Backbones}
We evaluate our GCNet on stronger backbones, by replacing ResNet-50 with ResNet-101 and ResNeXt-101 \cite{xie2017resnext}, adding deformable convolution to multiple layers (c3+c4+c5) \cite{dai2017dcnv1,zhu2018dcnv2} and adopting the Cascade strategy \cite{cai2018cascade}.
The results of our GCNet with GC blocks integrated in all layers (c3+c4+c5) with bottleneck ratios of 4 and 16 are reported.
Table \ref{table:archs-coco}(a) presents detailed results on the validation set.
It is worth noting that even when adopting stronger backbones, the gain of GCNet compared to the baseline is still significant, which demonstrates that our GC block with global context modeling is complementary to the capacity of current models.
For the strongest backbone, with deformable convolution and cascade RCNN in ResNeXt-101, our GC block can still boost performance by 0.8\%$\uparrow$ on AP$^\text{bbox}$ and 0.5\%$\uparrow$ on AP$^\text{mask}$.
To further evaluate our proposed method, the results on the test-dev set are also reported, shown in Table \ref{table:archs-coco}(b). 
On test-dev, strong baselines are also boosted by large margins by adding GC blocks, which is consistent with the results on the validation set. 
These results demonstrate the robustness of our proposed method.

\subsection{Image Classification on ImageNet}
ImageNet \cite{deng2009imagenet} is a benchmark dataset for image classification, containing 1.28M training images and 50K validation images from 1000 classes. We follow the standard setting in \cite{he2015resnet} to train deep networks on the training set and report the single-crop top-1 and the top-5 errors on the validation set.
Our preprocessing and augmentation strategy follows the baseline proposed in \cite{xie2018bag} and \cite{hu2018senet}.
Concretely, the following augmentation and preprocessing are performed sequentially during training: 
$[-10\degree, 10\degree]$ random rotation, 
$[3/4, 4/3]$ random aspect ratio with $[8\%, 100\%]$ random area crop, $224\times224$ resizing, 
horizontally flip with 0.5 probability, 
$[0.6, 1.4]$ HSV random scaling, and
PCA noise sampled from $\mathcal{N}(0, 0.1)$. 
The standard ResNet-50 is trained for 120 epochs on 4 GPUs with 64 images per GPU (effective batch size of 256) with synchronous SGD of momentum 0.9. 
Cosine learning rate decay is adopted with an initial learning rate of 0.1. 
\begin{table}[]
    \small
    \centering
    \addtolength{\tabcolsep}{-2.5pt}
\begin{tabular}{c|cc|c|c}
\Xhline{1.0pt}
\multicolumn{5}{c}{(a) \textbf{Block Design}}                                       \\
 & Top-1 Acc & Top-5 Acc & \#params(M) & FLOPs(G) \\
\hline
baseline & 76.51 & 93.35 & 25.56 & 3.86 \\
+1NL & 77.21 & 93.64 & 27.66 & 4.11 \\
+1SNL & 77.10 & 93.56 & 26.61 & 3.86 \\
+1GC & 77.20 & 93.47 & 25.69 & 3.86 \\
+all GC & \textbf{77.49} & \textbf{93.67} & 28.08 & 3.87 \\
\Xhline{1.0pt}
\multicolumn{5}{c}{(b) \textbf{Pooling and fusion}}                                       \\
 & Top-1 Acc & Top-5 Acc & \#params(M) & FLOPs(G) \\
\hline
baseline & 76.51 & 93.35 & 25.56 & 3.86 \\
avg+scale & 77.14 & 93.57 & 28.07 & 3.87 \\
avg+add & 77.16 & 93.63 & 28.07 & 3.87 \\
att+scale & 77.18 & 93.58 & 28.08 & 3.87 \\
att+add & \textbf{77.49} & \textbf{93.67} & 28.08 & 3.87 \\
\Xhline{1.0pt}
\end{tabular}
    \caption{\textbf{Ablation study} of GCNet with ResNet-50 for \textbf{image classification} on {ImageNet} validation set.}
	\label{table:ablation-imagenet}
\end{table}

\begin{table}[]
    \small
    \centering
    \addtolength{\tabcolsep}{-2.5pt}
\begin{tabular}{c|cc|c|c}
\Xhline{1.0pt}
 & Top-1 Acc & Top-5 Acc & \#params(M) & FLOPs(G) \\
\hline
baseline & 76.51 & 93.35 & 25.56 & 3.86 \\
SENet$^{*}$\cite{hu2018senet} & 76.86 & 93.30 & 28.07 & 3.87 \\
CBAM$^{*}$\cite{woo2018cbam} & 77.34 & \yue{\textbf{93.69}} & 28.07 & 3.87 \\
GCNet & \textbf{77.49} & 93.67 & 28.08 & 3.87 \\
\Xhline{1.0pt}
\end{tabular}
    \caption{\jiarui{Comparison of state-of-the-art methods with ResNet-50 for \textbf{image classification} on {ImageNet} validation set. * denotes that the results are directly taken from the original paper.}}
	\label{table:sota-imagenet}
\end{table}

\textbf{Block Design.}
As done for the block design on COCO, results on different blocks are reported in Table \ref{table:ablation-imagenet}(a). 
The GC block performs slightly better than the NL and SNL blocks with fewer parameters and less computation, which indicates the versatility and generalization ability of our design. 
By inserting GC blocks in all residual blocks (c3+c4+c5), the performance is further boosted (by 0.98\%$\uparrow$ on top-1 accuracy compared to baseline) with marginal computational overhead (0.26\% relative increase on FLOPs). 
\jiarui{In comparison to the baseline, a GC block requires about $250\times$ less computation than an NL block, i.e. 0.001G vs. 0.25G, which is significant. }

\textbf{Pooling and fusion.}
The functionality of different pooling and fusion methods is also investigated on image classification. Comparing Table \ref{table:ablation-imagenet}(b) with Table \ref{table:ablation-coco}(f), it is seen that attention pooling has greater effect in image classification, which could be one of the missing ingredients in \cite{hu2018senet}. 
\jiarui{Also, attention pooling with addition (GCNet) outperforms vanilla average pooling with scaling (SENet with layer norm) by 0.35\% on top-1 accuracy with almost the same \#params and FLOPs.} 

\jiarui{\textbf{Comparison with Other Approaches.} As shown in Table \ref{table:sota-imagenet}, we compare our approach with other state-of-the-art approaches on image recognition of ImageNet, and find that our GCNet outperforms SENet \cite{hu2018senet} and CBAM \cite{woo2018cbam}.}

\begin{table}[]
    \small
    \centering
    \addtolength{\tabcolsep}{-2.5pt}
\begin{tabular}{c|cc|c|c}
\Xhline{1.0pt}
method & Top-1 Acc & Top-5 Acc & \#params(M) & FLOPs(G)\\
\hline
baseline & 74.94 & 91.90 & 32.45 & 39.29 \\
\hline
+5 NL    & 75.95 & 92.29 & 39.81 & 59.60 \\
+5 SNL   & 75.76 & \textbf{92.44} & 36.13 & 39.32 \\
+5 GC    & 75.85 & 92.25 & 34.30 & 39.31\\
+all GC  & \textbf{76.00} & 92.34 & 42.45 & 39.35\\
\Xhline{1.0pt}
\end{tabular}
    \caption{Results of GCNet and NLNet based on Slow-only baseline using R50 as backbone on \textbf{Kinetics} validation set.}
	\label{table:ablation-kinetics}
\end{table}

\begin{table}[]
    \small
    \centering
    \addtolength{\tabcolsep}{-2.5pt}
\begin{tabular}{c|cc|c|c}
\Xhline{1.0pt}
method & Top-1 Acc & Top-5 Acc & \#params(M) & FLOPs(G)\\
\hline
Slow-only \cite{feichtenhofer2018slowfast} & 74.94 & 91.90 & 32.45 & 39.29 \\
GloRE$^{*}$ \cite{chen2019graph} & 75.12 & - & - & 28.90 \\
NLNet \cite{wang2017non}    & 75.95 & 92.29 & 39.81 & 59.60 \\
GCNet  & \textbf{76.00} & \yue{\textbf{92.34}} & 42.45 & 39.35\\
\Xhline{1.0pt}
\end{tabular}
    \caption{\jiarui{Comparison of state-of-the-art methods with R50 as backbone on \textbf{Kinetics} validation set. * denotes that the results are directly taken from the original paper. The GloRE results are based on lite version of C2D, and thus have lower FLOPs.}}
	\label{table:system-kinetics}
\end{table}
\begin{figure*}[!htb]
    \centering
    \includegraphics[width=1.0\linewidth]{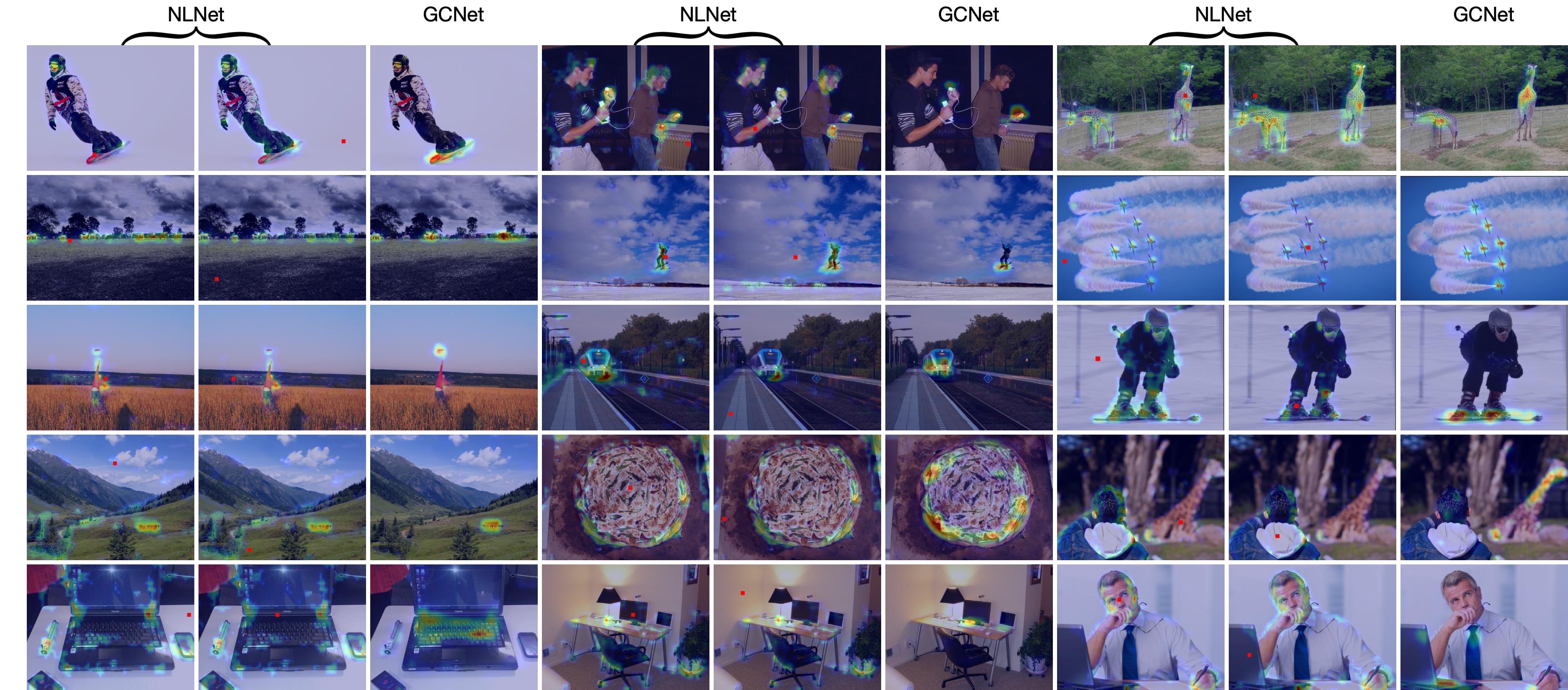}
	\caption{\yue{Visualizations of context modeling attention maps (heatmaps) of GCNet and NLNet (red points denote query positions). Their learnt attention maps are mostly similar. Also, they learn to focus more on hard cases like relatively small size, deformation, occlusion, and blur. 
	\emph{Best viewed in color}.}}
	\label{fig:vis-context}
\end{figure*}
\begin{figure*}[!htb]
    \centering
    \includegraphics[width=1.0\linewidth]{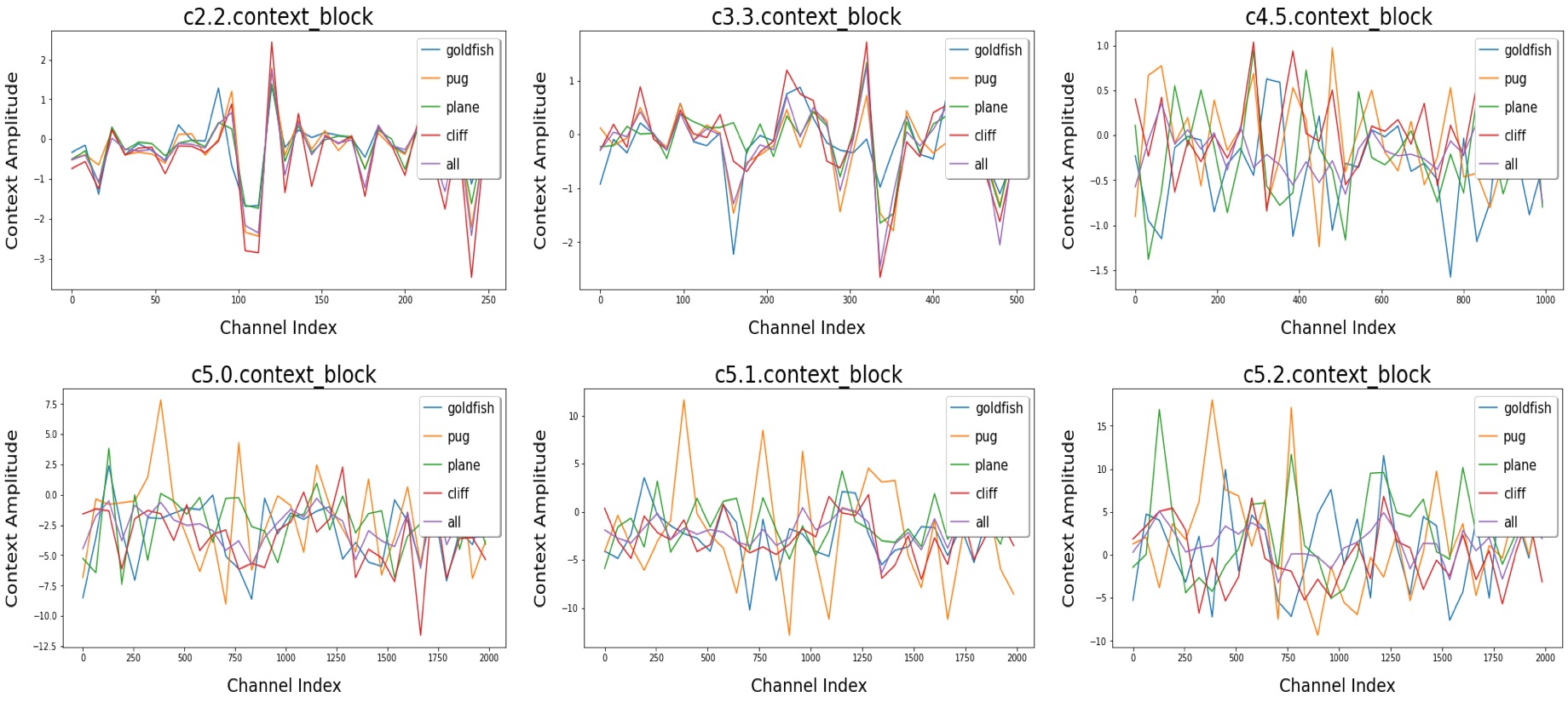}
	\caption{\jiarui{Activation of output of the transform function at different stages of GCNet. 
	\emph{Best viewed in color}.}}
	\label{fig:vis-activation}
\end{figure*}

\subsection{Action Recognition on Kinetics}
For human action recognition, we adopt the widely-used Kinetics \cite{kay2017kinetics} dataset, which has $\sim$240k training videos and 20k validation videos in 400 human action categories. 
All models are trained on the training set and tested on the validation set. 
Following \cite{wang2017non}, we report top-1 and top-5 recognition accuracy.
We adopt the slow-only baseline in \cite{feichtenhofer2018slowfast}, the best single model to date that can utilize weights inflated \cite{carreira2017i3d} from the ImageNet pretrained model.
The inflated 3D strategy \cite{wang2017non} greatly speeds up convergence compared to training from scratch. 
All the experiment settings follow \cite{feichtenhofer2018slowfast}; the slow-only baseline is trained with 8 frames ($8\times8$) as input, and multi(30)-clip validation is adopted.

\textbf{Ablation Study.} The ablation study results are reported in Table \ref{table:ablation-kinetics}. 
For the Kinetics experiments, the ratio of GC blocks is set to 4. 
First, when replacing the NL block with the simplified NL block and GC block, the performance can be regarded as on par (0.19\%$\downarrow$ and 0.11\%$\downarrow$ in top-1 accuracy, 0.15\%$\uparrow$ and 0.14\%$\uparrow$ in top-5 accuracy). 
As in COCO and ImageNet, adding more GC blocks further improves results and outperforms NL blocks with much less computation.

\jiarui{\textbf{Comparison with Other Approaches.} As shown in Table \ref{table:system-kinetics}, we compare our approach with other state-of-the-art action recognition methods on Kinetics, and find that our GCNet outperforms GloRE \cite{chen2019graph} and NLNet \cite{wang2017non}. }

\subsection{Semantic Segmentation on Cityscapes}
The Cityscapes\cite{cordts2016cityscapes} dataset is one of the most popular benchmarks for semantic segmentation, 
consisting of 5,000 high quality pixel-level finely annotated images and 20,000 coarsely annotated images captured from 50 different cities.
Only the finely annotated part of the dataset is utilized in our experiments, 
and is divided into 2,975/500/1,525 for training, validation and testing. 
In total there are 30 semantic classes provided, 19 of which are used for evaluation. 
The standard mean Intersect over Union (mIoU) on the validation set is reported for measuring segmentation accuracy.

The training setting and hyper-parameters strictly follow CCNet\cite{huang2018ccnet}. 
The data are augmented by random scaling the original $2048 \times 1014$ high resolution images by a factor in $[0.5, 2]$, then randomly cropping to $769 \times 769$ patches. 
The poly learning policy is employed where the initial learning rate 0.01 is multiplied by $(1 - \frac{\text{iter}}{\text{iter}_{\text{max}}})^{0.9}$. 
SGD training is performed on 4 GPUs with 2 images per GPU with Synchronized Batch Normalization for 160 epochs, which is roughly 60k steps. 
Following the practice of recent semantic segmentation approaches \cite{huang2018ccnet, chen2017rethinking, fu2019danet, yuan2018ocnet, zhao2017pspnet}, 
ResNet-101 pretrained by \cite{zhou2017scene, zhou2018semantic} is used as the backbone, where the downsampling operation in c4,c5 is removed and dilated convolution \cite{yu2015dilation} is incorporated. 
\yue{The backbone is followed by a semantic segmentation head. Like the design in CCNet \cite{huang2018ccnet}, the c5 feature is encoded by a context operator (e.g. CCNet, GCNet, SNLNet, NLNet) and concatenated with c5 before the pixel-wise classification layer. In the FCN \cite{long2015fully} baseline, there is no context operator. }
As done in previous works~\cite{zhao2017pspnet, yuan2018ocnet, huang2018ccnet, fu2019danet}, an auxiliary head is added after the c4 stage output for a deep supervision loss. 
We use ratio $r=4$ for the GC block as default for semantic segmentation experiments. 

\textbf{Block Design.}
As shown in Table \ref{table:ablation-cityscapes}(a), the SNL head achieves performance comparable to the NL Head. 
Hence we argue that the accuracy gains by self-attention can be mainly ascribed the modeling of global context rather than the learning of pairwise relations.
Moreover, all heads significantly boost the performance over the baseline, which indicates that long-range dependency is essential in the fine-grained semantic segmentation task. 
Note that with the GC block incorporated in the head, the GC blocks in the backbone do not have a significant effect because long range dependency is already exploited.

\textbf{Pooling and Fusion.} 
The observations for the pooling and fusion in \ref{table:ablation-cityscapes}(b) are similar to those of object detection. 
\jiarui{Moreover, attention pooling with addition (GCNet) outperforms vanilla average pooling with scaling (SENet with layer norm) with almost the same \#params and FLOPs.} 
We conjecture that simply recalibrating channels does not effectively exploit rich semantic global context. 

\begin{figure*}[!htb]
    \centering
    \includegraphics[width=1.0\linewidth]{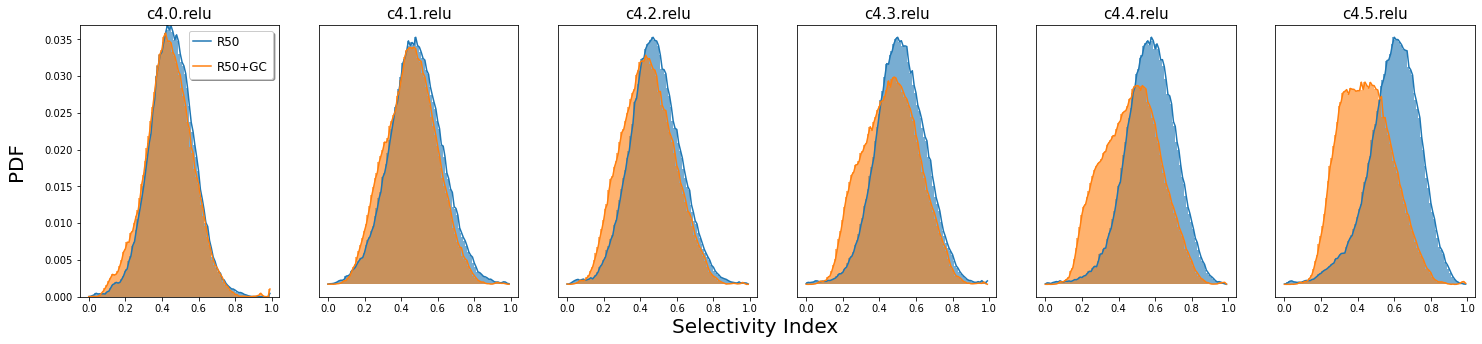}
	\caption{\jiarui{Distributions of class selectivity index of ResNet-50 baseline and ResNet-50+GCNet on different layers. For deeper layers, GCNet exhibits less class selectivity compared to the baseline. 
	\emph{Best viewed in color}.}}
	\label{fig:vis-class-selectivity}
\end{figure*}

\jiarui{\textbf{Comparison with Other Approaches.} As shown in Table \ref{table:system-cityscapes}, we compare our approach with other state-of-the-art approaches on semantic segmentation of Cityscapes, and find that our GCNet achieves performance on par with DANet \cite{fu2019danet}, ANN \cite{zhu2019ann}, CCNet \cite{huang2018ccnet} and NLNet \cite{wang2017non}. }

\begin{table}[]
\small
\centering
    \addtolength{\tabcolsep}{-1.5pt}
\begin{tabular}{c|c|c|c}
\Xhline{1.0pt}
\multicolumn{4}{c}{\textbf{(a) Block Design}} \\
 & \#params(M) & FLOPs(G) & mIoU(\%) \\
 \hline
baseline & 70.96 & 646.88 & 75.42 \\
NL Head & 71.22 & \yue{697.16} & 77.59 \\
SNL Head & 71.22 & 646.89 & 77.22 \\
GC head & 71.09 & 646.89 & 78.55 \\
GC backbone (c3-c5) & 89.92 & 647.19 & 78.49 \\
GC backbone + GC Head & 90.05 & 647.20 & \textbf{78.67} \\
\Xhline{1.0pt}
\multicolumn{4}{c}{\textbf{(b) Pooling and Fusion}} \\
 & \#params(M) & FLOPs(G) & mIoU(\%) \\
 \hline
baseline & 70.96 & 646.88 & 75.42 \\
avg+scale & 71.09 & 646.89 & 76.86 \\
avg+add & 71.09 & 646.89 & 77.84 \\
att+scale & 71.09 & 646.89 & 77.49 \\
att+add & 71.09 & 646.89 & \textbf{78.55} \\
\Xhline{1.0pt}
\end{tabular}
\caption{\textbf{Ablation study} of GCNet with ResNet-101 on \textbf{semantic segmentation} on {Cityscapes} validation set.}
\label{table:ablation-cityscapes}
\end{table}

\begin{table}[]
\small
\centering
\begin{tabular}{c|c|c|c}
\Xhline{1.0pt}
 & \#params(M) & FLOPs(G) & mIoU(\%) \\
 \hline
baseline$^\dagger$ & 70.96 & 646.88 & 75.52 \\
DANet \cite{fu2019danet} & 71.29 & \yue{709.18} & \yue{\textbf{79.88}} \\
ANN \cite{zhu2019ann} & 67.89 & \yue{632.10} & 79.32 \\
CCNet$^\dagger$ \cite{huang2018ccnet} & 71.49 & \yue{653.52} & 78.90 \\ 
NLNet$^\dagger$ \cite{wang2017non} & 71.22 & \yue{697.16} & 78.57 \\
GCNet$^\dagger$ & 71.09 & 646.89 & 78.95 \\
\Xhline{1.0pt}
\end{tabular}
\caption{\jiarui{Comparison of state-of-the-art methods with ResNet-101 on \textbf{semantic segmentation} with stronger augmentation on {Cityscapes} validation set. \yue{The methods denoted with ``$^\dagger$'' marker produce the pixel-wise classification logits by the concatenation of the stride-8 c5 backbone and the context head followed by a 3$\times$3 convolution layer, while the others directly utilize the context head features without concatenating the 2048-dim c5 features.}}}
\vspace{-10pt}
\label{table:system-cityscapes}
\end{table}

\subsection{Visualizations}
\yue{\textbf{Visualizations of Context Attention Map.}
In Figure~\ref{fig:vis-context}, we randomly choose fifteen images from the COCO dataset and visualize their attention maps (softmax output of {context modeling} module) for GCNet and NLNet. We can observe that NLNet learns similar attention maps for different query points in most cases, which are also similar to the attention maps learnt by GCNet. In addition, we observe that the two models usually focus on small or thin objects like frisbee, skateboard, and snowboard. This may facilitate the detection of these objects, and the accuracy is hence boosted. Also note that the human body is an exception, which is less attended. We hypothesize the reason is because the person class is common enough in the COCO dataset and it is not hard to be detected.}

\jiarui{\textbf{Output Activations of GC Block.} We follow \cite{hu2018senet} to visualize the output activations of GC blocks in different layers. As depicted in Figure~\ref{fig:vis-activation}, the channel activations are class-agnostic in the shallow layers and more class-dependent in the deeper layers. It is intuitive since for neurons closer to the final classification layer, a higher correlation between activation and class label is expected. }

\jiarui{\textbf{Illustration of Class Selectivity.}
We use the \textit{class selectivity index} proposed in \cite{morcos2018importance} to study the effect of global context modeling on learned representations.
In Figure \ref{fig:vis-class-selectivity}, we plot the distribution of the class selectivity index on ImageNet. We use the last activation of each block in the c4 stage to compute the class selectivity index. The observed pattern is similar to that in GENet \cite{hu2018gather}. The distributions are almost the same in the first blocks. As the depth increases, GCNet begins to diverge from the baseline. And as shown in the last plot (c4.5.relu) in Figure~\ref{fig:vis-class-selectivity}, GCNet exhibits much less class selectivity. Also pointed out in \cite{hu2018gather}, we speculate that there are some cases that suffer from local ambiguity, which would push the baseline network to specialize some neurons to overcome it. Note that the global context computed by GCNet may avoid this burden thus resulting in less class selectivity.}

\section{Conclusion}
The long-range dependency modeling of non-local networks intends to model query-specific global context, but we have found empirically that it only models query-independent context on several important visual recognition tasks. Based on this, we simplify non-local networks and abstract this simplified version to a global context modeling framework. Then we propose a novel instantiation of this framework, the GC block, which is lightweight and can effectively model long-range dependency.
Our GCNet is constructed via applying GC blocks to multiple layers, which generally outperforms simplified NLNet on major benchmarks for various recognition tasks.

We have verified that the global context block can benefit multiple visual recognition tasks. In the future, the global context block may be extended to the generative models \cite{goodfellow2014generative,kingma2013auto}, graph learning models \cite{kipf2016semi,velivckovic2017GAT}, and self-supervised models \cite{wang2019learning}.

\ifCLASSOPTIONcaptionsoff
  \newpage
\fi

\bibliographystyle{IEEEtran}
\bibliography{GCNet}

\end{document}